# The brain as a probabilistic transducer: an evolutionarily plausible network architecture for knowledge representation, computation, and behavior

Joseph Y. Halpern[1] and Arnon Lotem[2]

1. Department of Computer Science, Cornell University   halpern@cs.cornell.edu
2. School of Zoology, Faculty of Life Sciences and Sagol School of Neuroscience, Tel-Aviv University  lotem@tauex.tau.ac.il

The two authors contributed equally to the paper

**Abstract**

We offer a general theoretical framework for brain and behavior that is evolutionarily and computationally plausible. The brain in our abstract model is a network of nodes and edges. Although it has some similarities to standard neural network models, as we show, there are some significant differences.  Both nodes and edges in our network have weights and activation levels. They act as probabilistic transducers that use a set of relatively simple rules to determine how activation levels and weights are affected by input, generate output, and affect each other. We show that these simple rules enable a learning process that allows the network to represent increasingly complex knowledge, and simultaneously to act as a computing device that facilitates planning, decision-making, and the execution of behavior. By specifying the innate (genetic) components of the network, we show how evolution could endow the network with initial adaptive rules and goals that are then enriched through learning. We demonstrate how the developing structure of the network (which determines what the brain can do and how well) is critically affected by the co-evolved coordination between the mechanisms affecting the distribution of data input and those determining the learning parameters (used in the programs run by nodes and edges). Finally, we consider how the model accounts for various findings in the field of learning and decision making, how it can address some challenging problems in mind and behavior, such as those related to setting goals and self-control, and how it can help understand some cognitive disorders.

# 1. Introduction

While there have been an enormous number of theoretical studies addressing different aspects of brain and behavior, how evolution can create and refine the mechanisms that construct and run an extremely complex biological object like the brain is still poorly understood. The reason for that may be partly related to our incomplete understanding of neurobiological mechanisms and their genetic underpinning. Yet, following Marr (Marr, 1982), who advocated a computational understanding prior to, or independent of a biological understanding, and in line with similar approaches in evolutionary biology, where theoretical understanding frequently preceded genetic or molecular understanding (e.g., Smith, 1982; Grafen, 1984; Stephens and Krebs, 1986), we believe that constructing a computational theory of brain evolution may be possible and productive. Our goal is to offer a framework that is based on mechanisms that are sufficiently simple to be neurobiologically and computationally plausible, while still being able to evolve to attain the complexity of a working brain. Such a theory, we believe, may help to demystify how the brain works and how it could have evolved. It is certainly not sufficiently detailed or complete to allow us to build a brain, just as the theories of evolution and Mendelian genetics do not suffice to build an organism. But in a similar manner, we seek a theory that can "decode the magic" of the brain, using a set of rules and processes that can in principle do what the brain does, and can conceivably be improved and optimized by natural selection to the level of complexity found in the brain. Having such a theory may help to accelerate the development of a more complete and fully worked-out theory. Our approach has some similarities with current work in neural networks, but, as we shall see, there are a number of significant differences.

We have already taken initial steps towards this goal. In earlier work, we proposed a high-level framework for modeling how the brain represents and stores information using a network, where the nodes and edges in the network had weights. We then showed how this representation could be used to capture some aspects of cognitive evolution (Lotem and Halpern, 2012; Lotem *et al.*, 2017); related work demonstrated how it can capture the gradual transition from simple associative learning to increasingly complex cognitive processes, including those required for language acquisition and problem solving (Kolodny, Edelman and Lotem, 2014, 2015a, 2015b; Kolodny, Lotem and Edelman, 2015). While suggestive, the framework did not explain how an autonomous agent could use the network to make decisions, plan, and carry out policies. Clearly, understanding how such processes can be performed is necessary for providing a more complete model of the brain. Our goal in this paper is to show how our earlier framework can be extended to do this in an evolutionarily plausible way, while adding minimal additional structure.

We view each node and edge of the network that we construct as a very simple computational device. Formally, it is a *probabilistic transducer*. We discuss probabilistic transducers in more detail in Section 2. For our purposes here, it suffices to think of a probabilistic transducer as a simple device that, at any time, is in some state; when it receives a signal, it changes state and outputs a signal, according to some rules, which we can think of as its program. The reason that we call the computing device a *probabilistic* transducer is that the state that the transducer transitions to and its output signal are not uniquely determined by the input signal and initial state. Rather, for each initial state and input signal, the transducer has a probability distribution over states and possible output signals.

Simple probabilistic transducers can be combined to form more complicated transducers, so that we can think of the brain itself as a probabilistic transducer that is the result of combining all the simple transducers that make up the network. In Sections 3 and 4, we show how the type of probabilistic transducer we have in mind can be constructed so that it represents learned information in a useful way. While the resulting transducer can be viewed as a neural network, it differs from current architectures of neural networks in significant ways, which will become clear in Section 3. In Section 3, we also show how the network can perform basic operations, like simple associations, data acquisition, and data segmentation. In the process, we consider some evolutionary issues, showing how complex networks could have evolved starting from simple transducers. We continue in Section 4, showing how the network can represent more sophisticated operations like representing and detecting similarities, representing hierarchies, and dealing with multiple representations of the same data item.

In Section 5, we show how the network can use the information that it represents to decide how to act, just using the same programs at the nodes and edges that were used to construct the network in the first place. Specifically, we consider how the network can do prediction, decide between competing courses of actions, and do planning. In Section 6, we briefly discuss how the network can deal with challenging problems such as setting goals and priorities, self-control, and understanding causality, as well as helping in understanding complex cognitive disorders such as autism. We do not pretend to come anywhere close to completely resolving these problems -- we leave this to future work – but we find the fact that our approach can provide useful insights into these issues encouraging. We conclude the paper with some further discussion in Section 7.

# 2. The Brain as a Probabilistic Transducer

In this section, we motivate the use of a probabilistic transducer as a formal model of the brain, define transducers and the compositions of transducers carefully, and explain how we view the brain as a network of transducers.

## 2.1 Why probabilistic transducers?

Although the notion of a transducer comes from the computer science literature, transducers are ubiquitous. They can capture many biological mechanisms (Stieve, 1983), including the operation of neuronal cells (Harmon, 1959). Despite their conceptual simplicity, transducers can model a wide range of mechanisms. For example, a thermostat, which is a relatively simple device, can be viewed as a transducer: its state might record the temperature (and perhaps the time and other features, depending on how sophisticated the thermostat is); the input signal might come from a human setting a temperature and from sensor; the output signal goes to the furnace and air conditioning systems (which themselves can be viewed as transducers). A car, although much more complex than a thermostat, can also be viewed as a transducer. Its state describes (among other things) whether it is on or off, and what speed it is going at, and the input signals include turning the steering wheel, stepping on the accelerator, or stepping on the brake. As the result of getting a signal, a car transitions from its current state to a new state, and possibly outputs a signal (e.g., reports its speed on the speedometer). Although it may be easiest to think of a thermostat as deterministic, a car is more likely to be probabilistic. Putting the same amount of pressure will not always result in the car accelerating by the same amount (even in situations where the road conditions are identical). Rather, there is a probability distribution over car speeds.

One important feature of a probabilistic transducer is that the composition of simple probabilistic transducers is itself a probabilistic transducer. We define the notion of composition formally below, but the intuition can already be understood by considering a car. A car can be viewed as the composition of many subsystems, like the braking system, the cooling system, the exhaust system, and so on. The output of one subsystem can be the input to another subsystem. Each of these subsystems can itself be viewed as a probabilistic transducer. Transducers can thus be useful for analyzing a complex system at different levels of abstraction. The biological brain, for example, is a highly complex system composed of a huge number of neurons, each of which can be viewed as a transducer.

While it is difficult to model the entire brain at the level of the many neurons that compose it, by thinking in terms probabilistic transducers, we can capture the behaviors of networks of neurons and of complex subsystems of the brain without specifying or knowing exactly how this behavior is implemented at the neuronal level. We can thus offer a plausible theory despite existing gaps in knowledge.

When viewing the brain as a probabilistic transducer, we assume that the input signal comes from some other system in the body; for example, it could be a signal from the eye as a result of some visual input, a signal from the ear due to some auditory input, or an "I'm hungry" signal from the stomach. We sometimes use the term *(input) data stream* for the sequence of input signals received and call an individual input a *data item*. We assume that the incoming data stream may be discrete or continuous. In the case of auditory inputs, these units could consist of phrases in some language, snippets of songs, the barking of a dog, or music playing at the background. In the case of visual inputs, the units could consist of sequences of visual scenes. The amount of data in these data streams is determined by the size of working memory, which itself is determined by the structure and the state of the network, as will be explained later. Whether a data item is "meaningful" is determined initially by frequency of observation. That is, a data item its meaningful if it recurs sufficiently often. Thus, we implicitly assume that things that are common in the environment and are therefore frequently observed, are somehow meaningful. Clearly, some rare inputs may also be meaningful, and the brain will need to recognize this. We return to this issue below. Data items that are meaningful are more likely to change the states (and subsequently the weights) of the nodes and edges of the network; in some cases, they also lead to a signal being output that goes from the brain to another system in the body; for example, a signal to the vocal tract to produce some sounds or a signal to an arm or leg resulting in movement.

The idea of representing the brain as a massively parallel computational device is certainly not new. The idea of representing that computing device as a network is also not new; for example, it is present in the work on neural networks. Indeed, even the idea of defining a network that is simultaneously a representation of information and the computational device acting on that information (and whatever input signals it gets) is not new; for example, it is implicit in early work on connectionist approaches (see, e.g., Feldman and Ballard, 1982; Hopfield, 1982; Nolfi, Parisi and Elman, 1994). That said, we believe that the particular way we have of putting things together is novel and useful; we hope that this point will become clearer as we give more details of our model. Our approach may also be viewed as providing a middle ground between formal cognitive models that lack neuronal-like mechanistic structure (e.g., (Griffiths *et al.*, 2010) and neural-net connectionist approaches (e.g., Feldman and Ballard, 1982; McClelland *et al.*, 2010)), which are highly mechanistic and powerful, but have their own limitations. Most work on neural networks focuses on trying to build useful networks that can carry out some tasks, not necessarily on modeling the brain. The need to obtain good performance by such neural networks typically results in developing structures and mechanisms that may depart from biological realism (Crick, 1989; Bartunov *et al.*, 2018; Lillicrap *et al.*, 2020). Finally, although evolutionary algorithms may be used to optimize and improve the performance of neural networks (e.g., (Nolfi, Parisi and Elman, 1994), and see (Stanley *et al.*, 2019) for a review), little attention (if any) is given to the question of how such structures and mechanisms could have evolved in the biological world. Indeed, if we consider recent work on neural networks, typically there is some objective function that the network is trying to optimize, and it does so using gradient descent. As we shall see, there is no analogue to the objective function in our approach, nor do we use anything like gradient descent. Moreover, we are proposing a general-purpose brain that can learn many things, and acts in addition to learning. So although, like many in the neural network world, our approach is inspired by the biological brain, the technical details of what we do are quite different from those in the neural network literature.

Another line of research that is related to our approach is *neuromorphic computing* (Monroe, 2014), an approach to designing computers based on biological principles (typically with the goal of minimizing power consumption). As in our approach, there are analogues to neurons, which fire when they are sufficiently excited (where the excitation level of a neuron depends on the signals that it receives from other neurons). Although in neuromorphic computing both memory and computing are viewed as being distributed among the neurons, there is not the same emphasis on the network representing the memory and being a computing device. Moreover, while neuromorphic computing is (by design) inspired by the brain, its primary goal is to build a more efficient computer, not to explain the brain and its evolution.

As we hope to demonstrate further below, we designed our approach to make it evolutionarily plausible. That is, we consider how small modifications of the model's basic elements or parameters can make a better brain and how such incremental changes can explain how complex brains could have evolved from simple brains. In terms of evolutionary theory, this is like asking how small mutations or genetic variations can improve the functioning of the brain in a way that contributes to Darwinian fitness. This is important not only for "decoding the magic" of brain evolution, but also as a test of biological realism; we cannot have biological realism without evolutionary plausibility. This is also where thinking in terms of probabilistic transducers that are based on relatively simple rules may be useful.

## 2.2 Networks of Probabilistic Transducers

We start with a minimal formal description of probabilistic transducers for readers who want to understand the formal details. Readers that find it too technical can skip these two paragraphs, jumping directly to the example that follow them, and should still be able to understand the main ideas in the rest of the paper reasonably well. A probabilistic transducer is characterized by a state space S, a set $\Sigma$ of input signals, a set $\Sigma'$ of output signals, and a *transition function* F mapping S x $\Sigma$ to $\Delta$(S x $\Sigma'$), the set of probability distributions over S x Σ'.  Thus, F gets as input a state s in S (the initial or current state of the transducer) and an input signal σ ∈ Σ, and returns a probability distribution over states and output signals.  For example, given a state s and input signal σ, with probability 1/3 the transducer will change the state to s1 and output σ1, and with probability 2/3 the transducer will change the state to s2 and output σ2.  Thus, formally, a transducer is given by a tuple (S, Σ, Σ', F).  We can think of the transition function as describing the program that the transducer runs.  While both the state space S and the sets Σ and Σ' of input and output signals in a transducer can, in principle, be infinite, we will take them to be finite here (and, in fact, quite small).

A key property of probabilistic transducers is that they can be composed.   The idea is that given probabilistic transducers T1 = (S1, Σ1, Σ1',F1) and T2 = (S2, Σ2, Σ2',F2), where Σ1' = Σ2, so that the output signals of the first transducer (T1) are the input signals of the second transducer (T2), we can compose the two transducers into one larger transducer, whose input signals come from Σ1 (just like those of T1) and whose output signals come from Σ2' (just like those of T2).  For reasons that will shortly become clear, the state space of the composed transducer is S1 x S2, so a state in the composed transducer has the form (s1, s2), where s1 is a state of T1, and s2 is a state of T2.  We denote the composed transducer as T1 ∘ T2.  So T1 ∘ T2 has the form (S1 x S2, Σ1, Σ2', F), where F works as follows: given a state (s1, s2) and an input signal σ1 in Σ1, we first apply F1 to (s1, σ1), and then apply F2 to (s2, σ1), where σ' is the output signal generated by F1.   More precisely, given an initial state (s1, s2) and an input signal σ1, the probability transitioning to state (s1', s2') and outputting σ2 is the sum, taken over all output signals σ1' in Σ1', of the probability (according to F1) of transitioning from (s1, σ1) to (s1', σ1') times the probability (according to F2) of transitioning from (s2, σ1') to (s2', σ2).

This, in fact, is exactly how we want to think of the brain.  It is a "large" probabilistic transducer that can be viewed as the result of composing a network of much simpler transducers. The relationship between nodes and edges in the network can be viewed as relationships between transducers: a node n can be viewed as a simple transducer whose output signals are fed as input signals to all the edges leading out of n. These edges can also be viewed as transducers and the output signals of the edge leading from n to n', for example, are fed as input signals to the node n', which can also be viewed as a transducer, and so on.

To avoid confusion, we should stress again that a network of probabilistic transducers in not equivalent to a network of neurons. Rather, it is a high-level description of a computational device.  Because it is a high-level description, there may be a number of ways of implementing it.  We believe (and try to justify the belief, to some extent) that one such implementation could use real neurons (which can also be viewed as probabilistic transducers).  In this implementation, each node or edge in our network would be associated with, or composed of, neuronal structures involving a range of biological processes.

As mentioned above, we assume that certain nodes in the network receive signals from outside the brain. They may be sensory nodes that receive visual, auditory, olfactory, or tactile signals from the outside world, or they may receive signals from organs outside the brain, such as the stomach. While we do not believe that, in general, input signals carry any information about their source, we do believe that, in some cases, a node may be able to distinguish edges, and so recognize that input signal σ1 came on edge 1, input signal σ2 came on edge 2, and so on.[1] So we assume when describing a transducer we assume that input signals have the form (σ1, σ2, …), where σ1 comes from source 1, signal 2 comes from source 2, and so on. That said, in many cases all that may matter about the input signal is the collection of inputs, and not their source. (Mathematically, that means that the input signal should be treated as a set {σ1, σ2, …}, rather than a tuple.)

While we expect that the basic probabilistic transducers composing the brain (the nodes and edges) are mainly similar and relatively simple, the network they form can produce highly complex behavior. This is in part due to the ability to compose them in different ways. The ability to compose transducers allows us to analyze a network of transducers at different levels of abstraction. When describing the programs that the transducers run, we will start by considering the "finest" level of abstraction, and focus on the programs used by (the transducers associated with) single nodes and single edges. That is what we do in Section 2.4, after discussing how we can view the brain as a network of transducers.

**2.3 Modeling the Brain as a Network of Transducers**

As discussed in earlier work (Lotem and Halpern, 2012; Kolodny, Edelman, and Lotem, 2015a; Lotem *et al.*, 2017), we characterize the information acquired by the brain using a *graph*, or network; that is, a collection of nodes and edges; in Section 3 we discuss in more detail how it is constructed. For now, it is useful to recall some basic definitions from graph theory. A *directed graph* (also called a *directed network*) is a collection of *nodes* and *(directed) edges* (also called *(directed) links*). The fact that the edges are directed means that they go from one node to another. We often write n→n' to denote the directed edge from node n to node n'. We can think of clusters of nodes as representing data, broadly construed: they may represent individual features, like a morpheme or an acoustic note, or more complex data units such as individual objects in the world (including people), the performance of physical actions by the body, collections of stimuli such as hierarchies of words or scenes; or even more abstract notions, such as fear. How clusters of nodes come to represent particular objects or concepts in the world will become clearer as we progress through Section 3. Assuming that this can be done, we can think of the graph as describing how various meaningful notions are connected to each other.

---

[1] We believe that this may be possible, for example, for nodes that are themselves composed of a group of nodes and edges, so that the inner structure of such nodes makes it possible to distinguish between two signals coming from different edges (as in the case of a network of highways where cities are nodes,; having their own network of local roads allows us to distinguish highways coming from different cities).

To view the brain as a probabilistic transducer (that is itself a composition of many transducers), we need to define a state space, input and output signals, and a transition function. We describe the transducer here, and then discuss how it can be implemented by an actual brain.

- The states in the state space describe the state of each component (node and edge) of the graph characterizing the brain's information, and how they are connected (i.e., which edges lead out of each node, and which nodes an edge connects). Thus, we can think of a state as describing the current state of the graph (i.e., network). Each node and edge has a *weight* and an *activation level*. The weight and activation level characterize the state of each node and edge. Intuitively, the weight of a node (or edge) is a measure of that node's (or edge's) resistance to decay, which means that it is also a measure of memory; as we shall see, the increase in the weight of nodes and edges is the way by which the network represents and thus remembers learned information. *Learning* in our network is thus represented by a change in the weight of the nodes and edges (this will become clearer as we describe it in more detail throughout the paper). The activation level of a node (edge) can be thought of as characterizing the extent to which the information carried by that node or edge is currently affecting the network; the higher a node or edge's activation level, the more responsive it is to signals. (We discuss the notions of decay and responsiveness in more detail below.) Note that these assumptions, as well the fact that both nodes and edges have weight and activation levels, already make our model different from typical neural networks, where only edges have a weight and there is no analogue of activation levels. (Neural networks have *activation functions*, but these just define how the output of a node is calculated from its inputs.) We can think of the network structure and the weights in our framework as a representation of the information acquired by the brain, while the activation level is more "ephemeral", and describes what the brain is currently responding to. It is perhaps tempting to compare weights and activation levels in our model with long-term and short-term (or working memory), and they are indeed similar, but we refrain from doing that at this stage, as these latter terms themselves are not well defined (we will discuss later the extent to which our model can capture these phenomena). We assume that both weights and activation levels are numbers taken from a bounded interval. This is biologically realistic and also turns out to be useful for a number of mechanisms, as we show later.
- For simplicity, we take the input and output signals of the transducer to numbers, both positive (1, 2, and 3) and negative (-1, -2, and -3). Numbers larger in absolute value denote stronger signals. We think of the negative signals as being inhibitory. (Of course, a biological brain may not produce be able to produce "negative signals", but these can be implemented as long as there is some way to view some signals as inhibitory.). A signal of 0 is interpreted as the absence of a signal; see below.

- The transition function is best understood by thinking of the transducer that represents the brain as the result of composing many simple transducers, one corresponding to each node and each link in the network. We expect that, in practice, there are relatively few different programs associated with nodes and edges. As we said, the state of a node or an edge consists of its weight, and its activation level. We explain the transition function of these simple transducers (the program for nodes and edges) in the next subsection.

Two additional features of our model that should be mentioned at this stage are the assumptions of continuous activity and of dynamic development. The first implies that the nodes and edges of the network run their programs continuously (i.e., without stopping, even if we take their activity to occur at discrete steps), and on similar time scales (as a result of the same biological processes taking place at all nodes and edges). The second important feature of the network is its dynamic development. That is, nodes and edges can be created and the network can grow through learning, implying that it is to some extent open-ended. As will be explained below (Section 3.1), although we believe that much of the physical structure of the nodes and the edges is already in place at birth, the nodes and edges become effective only as a result of receiving sufficient input to raise their weight above zero. Thus, we can view the network as growing and developing as more nodes and edges get a weight above 0.

**2.4 The program for nodes and edges**

We first describe at a high level how the program for nodes works. A node n can receive an input signal from each edge that leads to n. (Recall that edges are directed.) Thus, the size of the space of possible input signals at a node n depends on the number of edges leading to that node and the size of the space of possible output signals of a node depends on the number of edges going from that node to other nodes. We assume that each node and edge has an internal clock that "ticks" at discrete time steps. (Our assumption that the nodes and edges run at similar time scales means that these clocks run at roughly the same rate, presumably as a result of being based on the same biological processes.) At each time step on its internal clock, a node may receive an input signal along an edge or not. If it receives no input signals, we view that as getting a signal of 0; in that case, both its weight and level of activation decrease by an amount that is a function of its current state. We can think of this as *decay.* There are two separate decay functions, one for weight and another one for activation level. For the weight, we also assume that there is a *fixation threshold*; once the weight reaches this threshold, it does not go below the threshold, even if the node receives few signals. If the weight is below the threshold, then the weight decays (i.e., decreases) in the absence of input; the lower the weight, the greater the decay. The exact quantitative details of the decay are not relevant at this point.

Recall that weight represents resistance to decay, which can be viewed as resistance to being forgotten. When the weight of a node decays to zero, we can view the (information represented by the) node as having been forgotten. (There may be additional mechanisms that result in forgetting; see Section 5). The assumption that the weight decays quickly once it is already low but, once it reaches a certain threshold, does not go below it (so that the rate of decay of weight is determined by the weight) is consistent with empirical observations: new information may be forgotten quickly unless it is repeatedly encountered, in which case it is remembered for a long time (Smith, 1968; Divjak and Caldwell-Harris, 2019). In earlier work (Goldstein *et al.*, 2010; Lotem and Halpern, 2012; Lotem *et al.*, 2017), we suggested that this form of memory decay is adaptive, as it provides an important test for the statistical significance of patterns in time and space (see also (Anderson and Schooler, 1991, 2000) for a similar intuition).

Since weights increase with repeated observations, we would like to view the relative weights of nodes as a proxy for the relative likelihood of the events represented by these nodes (i.e., a proxy for the relative frequency of these events in nature). However, if we assume that weights are bounded and that sufficiently small differences between weights are effectively indistinguishable (both of these assumptions seem reasonable in practice), then it seems that we have a problem. Suppose that the event associated with node n is observed twice as often as that associated with node n', but both of these events are observed frequently. Then the weights of both n and n' would both increase to the limit, and we would not be able to use weight as a proxy for n being twice as likely as n'.

We suggest here two quite different (and possibly complementary) ways by which the brain might handle this issue. The first is to assume that weights above the threshold are reset every so often (perhaps while the agent is sleeping), in such a way that they stay above threshold, but remain far enough away from the upper bound that we can allow for meaningful differences between weights to arise at all times. For example, the reset can cut the amount that the weight is above threshold by a certain factor, such as 20% (so that if the threshold is *t*, then the weight $w > t$ would be reset to $w - (w-t)*0.8$. Note that if such a reset happens every night, for example, then the weight would measure both statistical frequency and recency.

The second approach does not so much deal with the problem, but postpones it. The idea is that, rather than always increasing the weight by the same constant every time an observation is made, we lower the constant the higher the weight is, so that we might initially increase by $c_1$, then by $c_2 < c_1$, then by $c_3 < c_2$, and so on. As before, at some point we still hit the threshold, but now (with the appropriate choices of increase), that point comes much later. Moreover, the brain can still use weight as a proxy for relative likelihood up to that point.

Both approaches seem plausible to us and may even be combined; there are surely other plausible approaches as well. Resolving this issue, and understanding how the brain actually handles it, is beyond the scope of this paper. What matters here is that there are reasonable ways that we can assign weights to nodes such that we can take weight to be a reasonable (albeit imperfect) proxy for relative frequency, while at the same time assuming that the weight does not go below a certain threshold once it is above it.

Note that, in contrast to the weight, which we assume changes relatively slowly once it is sufficiently high, these issues do not arise for level of activation, since it is more ephemeral and it changes more rapidly in response to the actual situation than the weight does.

If a node receives input signals along some links, its weight and activation level change as a function of the strength of these signals, the number of signals it gets, and its current state (i.e., its current weight and activation level). Although our framework allows for the program at a node (which governs the change in weight and activation level) to be arbitrary, we expect that, typically, the greater the strength of the incoming signals, the greater the increase in weight and activation of the node. If the signals are weak or negative (recall we view negative signals as inhibitory), the activation level and/or weight may decrease. As we said, we identify not getting a signal with getting a signal of strength 0; thus, the decay of the weight and activation level as a consequence of not getting a signal can be viewed as a special case. The details of the functional form of the change in weight and activation level are not relevant at this point. When we speak of "typical nodes", we mean nodes whose program works this way. However, not all nodes are necessarily typical nodes in this sense.

Finally, a node sends output signals to all edges leading out of the node with a probability that increases with the increase in its level of activation. The output signal from a node to an edge is determined by the activation level of the node, the weight of the node, and the strength of the signal. In the typical case, a higher activation level, a greater weight, and a stronger input signal all result in a stronger output signal. This is the sense in which we view the activation level as measuring responsiveness to signals. The output signal of a node may be stronger than that of any of its input signals; for example, getting three moderately strong input signals may result in a much stronger output signal. However, we stress that what happens here is probabilistic; the same input may not always result in the same output, even if the node has the same weight and activation level. So, more precisely, we should say that, in the typical case, greater weight, higher activation levels, and stronger input signals all lead to a higher probability of a stronger output signal.

The transition function of the transducer associated with an edge runs a program that is essentially the same as that associated with a node. While a node can receive signals from multiple edges (all the edges that lead to that node), since edges are directed, an edge can receive a signal from only a single node: the one at its tail. Again, at each time step that no signal is received, both the weight and level of activation of the edge decay, just as they do for a node. And when an edge receives an input signal, we expect that both its weight and activation level will increase as a function of the strength of this signal and its current state (i.e., the current weight of the edge and its current level of activation). However, as in the case of nodes, there might be some atypical edges for which this does not happen. Finally, the edge sends an output signal to the node at its head. And just as in the case of nodes, typically, the strength of the output signal is a function of the increase in activation level of that edge (that, in turn, is a function of the input signal strength, the edge's weight, and its activation level).

# 3. How the Network is Constructed: Simple Associations, Data Acquisition, and Data Segmentation

Our notion of a network with nodes and edges is designed to be an abstraction of a number of commonly occurring scenarios. The same network can be implemented in many different ways. Intuitively, we would expect a node in our network to be represented by a neuron or group of neurons in the brain and the edges to be represented by synaptic structures. But identifying edges with synapses might be too simplistic; edges might also be represented by, for example, the secretion of neuromodulators such as dopamine, which provide a way for one neuron to regulate the behavior of other neurons. One advantage of thinking in terms of networks is that we can abstract away from the details of how nodes and edges are implemented. But it should be clear that it is possible to implement a network of the form that we have in mind using objects that we actually see in the brain, and that these objects can also implement the programs that we envision being run by nodes and edges.

Of course, it is not enough just to know that a network of the form that we have in mind can be implemented. Much more needs to be done, including the following:

- We need to show how the network can be constructed so that it represents learned information in a useful way. As mentioned earlier, we have already outlined a basic theory for doing this in earlier work (Lotem and Halpern, 2012; Goldstein *et al.*, 2010; Kolodny, Edelman, and Lotem, 2015a, 2015b; Kolodny, Lotem, and Edelman, 2015; Lotem *et al.*, 2017). Our goal here is to advance this theory by describing it in terms of probabilistic transducers.

- We need to explain how the ability of nodes and edges to construct a useful network could have evolved.
- We need to show that the network can behave (i.e., act and compute) as we would expect the brain of an autonomous organism to behave.

In this section, we take a first step at addressing the first issue, by showing how the network can perform basic operations, like simple associations, data acquisition, and data segmentation. In the process, we will need to consider evolutionary issues as well. In Section 4, we continue addressing the first issue, by considering further issues of knowledge representation, like the detection of similarities in the data, which will allow us to deal with multiple representations of the same object, hierarchies, and context. In Section 5, we address the final issue of how the network can act and compute as we would expect the brain of humans and other animals to behave.

## 3.1 The Development and Evolution of the Network

In order for us to examine how the network is able to perform basic operations such as simple associations, it is useful to think about how the network starts and what allows it to develop properly. It is convenient for us to think of the network as existing in the brain from birth (or even before birth, as some learning may start in the embryonic stages). That is, the nodes and the edges are in place even in the embryonic stage.[2] We expect that, initially, the weight of most nodes and edges in the network is zero. A weight of zero does not mean that the nodes and edges do not exist, but that their weight is at some baseline level. A network whose nodes and edges have weight zero is pretty much like a blank slate. However, we do not want to assume a complete blank slate! Not all weights are zero. We assume that some sections of the network has nodes and edges whose weights are above zero right from the start. This is how innate information is encoded in the network. Note that only the sections of the network with weight significantly more than 0 are likely to transmit signals effectively. This means that only innate behaviors can be produced by the network at this stage.

It is critical in our view (see also (Elman, Bates and Johnson, 1996)) that the development of the network arises through the interplay of the data observed, the innate sections of the network, and the programs of the nodes and edges, all of which have co-evolved through natural selection. This is important for several reasons, some of which will become clear only towards the end of this section. Among the many uses of the network, three particularly important ones are to decide if novel data inputs are relevant for the organism, to classify various types of input as "good" or "bad" in some sense, and to associate relevant inputs with appropriate adaptive responses. Food, for example, should be eaten; a potential mate should be courted (rather than being eaten); and predators or toxins should be avoided. In other words, incoming data should be matched with what biologists and psychologists view as "innate templates", "reinforcers", or "emotional systems" that elicit appropriate responses and allow the incoming data to be associated with the relevant context (Shettleworth, 2009). In our model, this is captured by assuming that some sections of the network are innately programmed to respond to specific input in specific ways. For example, some sensory nodes or some specific areas of the network (programed to respond to specific patterns of sensory input) are innately associated with pleasure, pain, fear, sexual arousal, and so on, and these innate "reinforcers" or "reward nodes" are already associated with the sections of the network that can initiate appropriate actions. Formally, this says that both the programs of the transducers as well as the initial (innate) states of the transducers have been shaped by evolution.

---

[2] Additional physical structures may be recruited later in life to enable further growth of the network. Plastic changes of this kind, such as the production of new neuronal cells and their recruitment to specific areas of the brain, are well documented (e.g. (Paton and Nottebohm, 1984)). For simplicity, we ignore this process here.

The assumption that the network has such innate components is critical for making our model evolutionarily plausible: First, it provides the genetic variation on which selection can operate (see (Lotem and Halpern, 2012; Lotem *et al.*, 2017) and further discussion below). Second, it may help in understanding how complex brain networks could have evolved. Initially, in primitive organisms, behavior was generated by innate stimulus-response rules that can be modeled as a set of probabilistic transducers. The "brain" of such simple organisms can be viewed as a probabilistic transducer: it is composed of all the stimulus-response transducers that control the organism's behavior. Note, however, that the evolution of associative learning (the mechanism of associating two stimuli in memory representation that will soon be addressed in more detail) and the ability to construct complex associative networks, may be viewed, at least initially, as nothing more than an extension and elaboration of these innate rules. The representation of learned information is useful because it allows innate responses to be conditioned on experience, and thus to expand the circumstances under which adaptive responses can be generated. For example, after some training, an innate response to a particular type of food can be triggered by a range of novel stimuli that were associated with this type of food in the past, and a search for this food can be carried out by navigating the network that represents how these different stimuli are associated to this food in time and space (see, e.g., (Kolodny, Edelman, and Lotem, 2015a)). Note that the stimulus of food also acts as a reinforcer. It reinforces the association between various stimuli and food or the action of responding to food. As we explain in some detail below, while the innate rules can be captured by innate sections of the network, the experience associated with these innate rules is represented by parts of the network that are constructed during the learning process. Both the innate and the learned parts of the network have the same "design": they are composed of nodes and edges that can be viewed as probabilistic transducers running essentially the same program.

The upshot of all this is that that the structure of the agent's initial network (i.e., those parts that have nonzero weight, the selection of stimuli that the nodes of nonzero weight respond to, and the action that they elicit) has a significant impact on what the agent pays attention to and on later development. (We explain these points in more detail in later sections.)

Natural selection is likely to affect the network in other ways as well. The brain must have some rules that determine how and when the weight and activation level of nodes and edges is increased and decreased, and what the fixation threshold is. These rules may be viewed as the parameters of our model.[3] We assume that these parameters are largely innate and can vary genetically, so that natural selection can select the parameters that result in constructing the network that maximizes fitness. But how can the network maximize Darwinian fitness? As we explain below, the relative weight of nodes and edges represents, in part, statistical information about the environment. A "good" network would therefore be a network that provides a useful and reasonably reliable representation of the environment. But even more important than modeling the environment well is to "act well"; that is, a good network is one that can be used to generate adaptive behavior (a behavior that increases the biological fitness of the animal). This is the ultimate test of the network.

Thus, natural selection results in better networks being constructed as a result of selecting the initial structure of the network, the weight increase and decrease parameters, and the fixation threshold. A "good" choice will lead to the construction of a good network, that is, one that provides useful representations that can be used by the animal to enhance its fitness.

Finally, just having the initial network and the network's parameters evolve may not be enough for improving fitness. The neuronal structures and brain circuits that realize the network are likely to have size and morphology constraints that are at least partly determined genetically. That is, adaptive changes in the programs of nodes and edges that can potentially lead to the construction of an extensive network in the acoustic domain, for example, may be subject to physical constraints that are in part genetically determined. We expect therefore that, over generations, genetic variants that are better at relaxing these physical constraints and in meeting the demand for larger or more appropriate neuronal structures will be favored by selection. This view of brain evolution is consistent with the "Baldwin effect" and the "genetic accommodation" views of evolution (Baldwin, 1896; Weber and Depew, 2003), according to which genes are selected based in part on how well they support adaptive plastic processes such as learning.

---

[3] Readers who are familiar with neural nets terminology may find it more useful to compare these parameters in our model to "hyper-parameters" in neural nets, that is, the parameters that are set before the learning process, such as the learning rates and the structure of the network, rather than the parameters that are adjusted through learning. (Note that in our model we do not view weight increase or decrease as an adjustment of the network parameters as in neural nets; the weight of a node or an edge in our model is part of its state.)

We can use this approach to address, for example, the question of how the cultural evolution of language or toolmaking shapes the evolution of the human brain. It implies that these cultural innovations exert selective pressure that shapes the programs of the nodes and edges that construct the network, promoting the construction of networks that are optimized given the physical constraints. Again, over evolutionary time scales, brain anatomy may be selected to better accommodate the physical requirements of a more adaptive network (see (Lotem *et al.*, 2017) for further discussion of these issues).

**3.2 Constructing simple associations**

Before we explain how the probabilistic transducers that form the nodes and edges of the network can construct complex networks (and can thus facilitate complex behaviors), we have to clarify how they can construct the kinds of simple associations that we find in basic forms of associative learning.

Let's start with a simple version of associative learning, of the type that allows animals to learn food-related cues. This may occur when the increase in the activation level of a node as a result of receiving sensory input of a novel stimulus coincides with the activation of a "reward node" as a result of sensing a reward within the same time frame (i.e., at the same time or shortly before or after). Assuming that these co-activated nodes can affect each other's activation levels through the edges and nodes connecting them in the network (see further details below), the high weight of the reward node (a node that represents, for example, the sweet taste of sugar, and innately has high weight) implies that the weight of other nodes and edges co-activated with it increases quite rapidly. This is because, in typical nodes, high weight facilitates activation, and higher activation causes a further increase in weight. We can also assume that because obtaining rewards increases fitness, the program of the reward nodes has evolved to respond to an input signal by sending a strong output signal, which is useful for two reasons: First, it increases the probability of executing adaptive actions (see Section 5 for more details). Second, sending a strong signal increases the activation level, and thus the weight of the nodes and edges connecting the reward node with both the novel stimuli and the appropriate responses to them; this is how the weight of these nodes and edges increases above zero and represents the learned association.

As there are many types of rewards (e.g., many different receptors responding to various types of food or aversive toxins), there are potentially many reward nodes receiving sensory input from such receptors, as well as from the visual and the auditory systems. While there is much ongoing research about the details of the reward system in the brain, in our framework its operation is basically like that of the simple reward nodes described above. That is, these are nodes that are associated with a specific set of pre-programmed functional responses (eat, avoid, etc.); due to their initial high weight, they can increase the probability of executing these responses in the future by increasing the weight of the nodes and edges that were co-activated with them in the past.

The description above already indicates that, according to our framework, we can think of many specific forms of learning (associative learning, reinforcement learning, etc.) as special cases of what is generally viewed as *Hebbian learning*. To capture this, we assume that if there is an edge in the network from n to n', there is also an edge going back from n' to n. We assume that the program of a typical node n' is such that when n' receives a signal on the edge from n to n', then it sends a (weaker) signal back on the edge from n' to n. Thus, a signal sent from n to n' is likely to cause a loop of activation between n and n' that increases the activation level of the edges from n to n' and from n' to n. Because the signal sent back is weaker than the original signal, this "activation loop" will decay quite rapidly. Nevertheless, it ensures that the weights of edges transmitting signals between n and n' increases. This is analogous to the strengthening of the connections between neurons that fire in sequence, as in Hebbian learning (Hebb, 1949), and captures the intuition that "cells that fire together wire together".[4]

Hebbian learning is often viewed as a form of unsupervised learning, while conditioning (classical or operant) represents some form of reinforcement learning that may be viewed as supervised. This is because the operation of the reward nodes has presumably evolved to reinforce behaviors that are evolutionarily "correct" (increase fitness) and thus to "supervise" the learning process. In our framework, both supervised and unsupervised learning are instances of nodes and edges running similar programs. However, supervised learning in our network typically requires fewer repeated co-activations than Hebbian-like learning, because the node representing a novel stimulus (or a novel action) is co-activated with the "reward node", which innately has high weight. With Hebbian-like learning, we typically need more repeated co-activations (i.e., repeated observations) because the initial weight of the nodes and edges that are involved is much lower than that of reward nodes. In fact, at the onset of the process, the weight of the nodes and the edge connecting them may be zero. Thus, it takes repeated activation loops to associate nodes through Hebbian-like learning in our network. As mentioned earlier, this is adaptive; we do not want the network to learn and represent any co-incidental association. Only repeated co-activations are likely to represent real associations; the need for repeated co-activation acts as a test of statistical significance (Goldstein *et al.*, 2010). Some repetitions may also be important for improving the reliability of reinforcement learning, but the association with a reward node already signifies potential importance; the greater the importance, the fewer the number of repetitions needed to learn. For example, in the case of danger, taste aversion, or other ecologically significant events, a single event may be sufficient (Garcia, Ervin and Koelling, 1966; Roozendaal, 2002; LaBar and Cabeza, 2006; Roberts *et al.*, 2012).

---

[4] Our model can also account for anti-Hebbian learining, where co-activation results in reducing associative strength, if we assume that signals are inhibitory (i.e., have negative values).

A topic closely related to associative learning is the notion of learning as a process that makes it possible to generate predictions and minimize *prediction error* (Rescorla and Wagner, 1972; Friston, 2010). We discuss how this can be done by our networks in Section 5.1.

### 3.3 Constructing a complex network

Scaling up from simple associations to complex network involves a few challenges. The model presented here and the way it handles these challenges has already been described in several previous papers, where further details can be found (Lotem and Halpern, 2008; Goldstein *et al.*, 2010; Kolodny, Edelman, and Lotem, 2015a, 2015b; Kolodny, Lotem, and Edelman, 2015; Lotem *et al.*, 2017). Some of the main aspects of the model were implemented in a set of computer simulations, demonstrating a gradual evolutionary trajectory, from simple associative learning, to chaining (second-order conditioning), to seldom-reinforced continuous learning, in which a network model of the environment is constructed (Kolodny, Edelman, and Lotem, 2014), to complex hierarchical sequential learning that can support advanced cognitive abilities of the kind needed for language acquisition and for creativity (Kolodny, Edelman, and Lotem, 2015a, 2015b). Our goal here is to explain how this computational model can produce many of the outcomes we expect of the brain using just the simple rules described above for the transducers at the nodes and edges.

Constructing a complex network that represents the environment requires processing the large amounts of data received continuously by the sensory system. Dealing with all this data requires deciding what to focus on, segmenting the data that is focused on into useful units (Gobet *et al.*, 2001; Jones, 2012), and constructing the network in a way that would facilitate efficient search, appropriate decision-making, and planning (Anderson *et al.*, 2004; Bellmund *et al.*, 2018). These issues are deeply intertwined. Note that there is a "bootstrapping" problem here: the current network must decide what to focus on and how to "grow" so as to construct a network that will prove useful in the future.

Some of these issues have certainly been considered before. For example, the field of statistical learning offers several methods to usefully segment data (e.g., (Brent, 1999; Solan, 2005; Conway, 2020)). Typically, these methods require much memory and computation, make use of entire data sets to detect statistical regularities, and rarely consider how evolution could endow the learning mechanism with prior knowledge that can improve and simplify its operation. Our approach suggests that the solution that has evolved to enable biological brains to handle these challenges uses selective data acquisition and limited memory ((Lotem and Halpern, 2012; Lotem *et al.*, 2017); see also (Perruchet and Vinter, 1998)). Our model is based on co-evolving mechanisms of learning and data acquisition. The learning mechanisms are those that give weight to nodes and edges; the data-acquisition mechanisms are those that determine the relevance and thus the distribution of data input (i.e., they determine what to pay attention to and what to ignore in the data stream). As illustrated below, the type of network that is constructed depends critically on the coordinated action of these learning and data-acquisition mechanisms.

We now briefly sketch how the network develops over time due to the joint effect of the learning and data-acquisition mechanisms, and how these mechanisms are related to the programs of nodes and edges. This involves describing how the network acquires data and "chunks" it into segments. The description we give in the next two subsections may become clearer and more intuitive after reading the description of the application of these principles to the simplified example outlined in Section 3.3.3.

**3.3.1** Data acquisition

The network changes over time. Initially, the network includes (a) the sensory nodes that determine what type of data can be sensed and (b) the innate sections of the network. An infant, or a young animal, uses the network for data acquisition. While it is easy to see that the sensory nodes have a major role in data acquisition, the role of the innate sections of the network requires explanation. To understand their role, we consider more generally the role of parts of the network that have higher weight (which is, by definition, the case for the innate sections in the initial network).

For simplicity, assume that the input to the learner takes the form of strings of data (i.e., linear sequences of discrete items). This assumption is reasonable for much linguistic input, which comes as a sequence of words (although the listener may not know where the word boundaries are, so what is heard is a sequence of morphemes). As we hope will be clear, the approach we suggest applies equally well to more complex data inputs. We expect our network to represent (some of) the data that it receives. We stress that all types of sensory data can be represented by nodes in the network. For example, there may be nodes that represent visual scenes (for sensory input arrives from the visual system), odors (for input that comes from odor receptors), and the perception of physical actions (if the input comes from motor neurons within the body). We say that the network *acquires* a string 's' of data if the patterns of sequential activation of sensory nodes associated with 's' are somehow represented by the nodes and edges in the network. As in the case of simple associative learning, this is done by increasing the weight of the nodes and edges that were activated by the sensory input. Thus, we can say that the increase in weight of the cluster of nodes and edges activated by 's' represents 's' in the network (we will see later that 's' may also be represented at a higher level by a single node that is uniquely activated by this cluster of nodes). Recall, however, that without repeated activation, the increase in weight of the cluster of nodes and edges representing 's' may decay quickly. In order to learn new data, we need some way to combat the decay. But we must be careful in how we do this. Decay is generally adaptive for filtering out non-significant data patterns (Lotem and Halpern, 2008; Goldstein *et al.*, 2010). We want to retain the data strings that contain important information yet are not observed sufficiently often to overcome the decay, while still allowing the "less useful" data strings to decay. This is where portions of the network that have high weight play an important role. Much like the case of associative learning discussed above, where the "reward" is an innate node that has high weight, any portion of the network that already has high weight can facilitate data acquisition. If the input of a new data string partially matches that of a portion of the network, for example, the new data string ABCD partially matches the portion BC that is already represented in the network with some positive weight, the activation level of that portion increases relatively more (because higher weight leads to stronger activation, according to the program at each node and edge). This stronger activation of BC results in a greater weight increase of the co-activated nodes and edges, including those activated by A and D, making the nodes and edges that represent the newly acquired data ABCD less likely to decay.

Consider a more detailed example: Suppose that a sensory input (in this case, a data string) sequentially activates a set of six nodes: $n_1$, $n_2$, $n_3$, $n_4$, $n_5$, and $n_6$. This would result in the activation of these nodes and the directed edges connecting them, that is, the subnetwork $n_1 \to n_2 \to n_3 \to n_4 \to n_5 \to n_6$, and in an increase of the weights of these nodes and edges. However, the activation and the higher weights may soon decay to zero. But if nodes $n_3$ and $n_4$ and the edge between them already have higher weight, then the activation level of the edge from $n_4$ to $n_5$ is higher than it would be if $n_4$ didn't have higher weight; this higher activation level is propagated to nodes $n_5$ and $n_6$ and the edge connecting them. Similarly, greater activation is propagated back to $n_2$ and $n_1$ by the "back-edges" corresponding to the edges from $n_1$ to $n_2$ and from $n_2$ to $n_3$. The greater activation level results in a significant increase in weights, thus increasing the probability that the representation of the entire data string will eventually reach fixation in the network. Thus, the greater weight of $n_3$ and $n_4$ can facilitate the acquisition of the data string that will now be represented by the subnetwork $n_1 \to n_2 \to n_3 \to n_4 \to n_5 \to n_6$. Clearly, as in the simple case of associative learning described earlier, the effect of the greater weight of $n_3$ and $n_4$ diminishes with the distance of the node. We would not expect it to facilitate the acquisition of a much longer data string (e.g., of a string that continues to $n_7 \to n_8 \to \ldots \to n_{15}$).

The upshot of this process is that the network will effectively "pay more attention" to strings parts of which already have relatively higher weight in the network. Thus, in our framework, what an agent is paying attention to is captured by (the information represented by) the parts of the network that have high activation level, and these are likely to be the parts that have higher weight. Moreover, the relevance of new data strings is determined by whether they contain segments that already have higher weight in the network and by their frequency of occurrence (since a sufficiently high frequency allows them to gain weight fast enough to prevent decay). Both of these conditions for "relevance" make ecological and evolutionary sense. Innate portions have presumably evolved in part based on how they facilitate the acquisition of useful data and the construction of an adaptive network, and patterns that occur with high frequency in the environment are likely to represent a significant aspect of the environment.

This discussion should also make the role of the innate network clear: it largely determines what the agent will pay attention to and acquire initially, and thus will affect the whole later construction of the network. This view of innateness is very different than the kind of nativism advocated by Chomsky (Chomsky, 1980), Pinker (Pinker, 2003) or Spelke (e.g., (Spelke and Kinzler, 2007), and is much more consistent with the developmental approach endorsed by Elman et al. (Elman, Bates and Johnson, 1996).

**3.3.2** Data acquisition and segmentation (chunking)

The process of data acquisition described so far is only the first step in constructing the network, but it is critical for subsequent stages. We previously suggested that data strings are processed in a working-memory buffer of relatively small size and tested for familiar segments and statistical regularities among their components (Lotem *et al.*, 2017). We now explain how the programs at each node and edge effectively ensure that this happens in the network.

The simplicity of our model requires that all cognitive terms and processes must be represented somehow in terms of nodes and edges and their weight and activation levels. This is also true for working memory. We view the nodes and edges activated by some input as the *working memory* of that input. Working memory is effectively created when sensory nodes that receive data input send signals that increase the activation level of some nodes and edges in the network. The duration of this working memory is defined by how long it takes before the activation of these nodes and edges decays to 0, and its size is determined by how many nodes and edges are affected by the sensory input. The level of activation of all nodes and edges in the working memory increases, so there is also an increase in the weight of these nodes and edges, giving rise to a dynamic learning process, as we now explain.

First, when a subsequence occurs several times within the longer sequence of data that is currently affecting the working memory, the nodes and edges responding to this subsequence are affected repeatedly. Thus, their activity and weight will increase relative to that of the other nodes affected by this data string (because these other nodes are affected only once). This greater increase in weight makes it more likely that repeated subsequences will be represented in the network. From our perspective, this subsequence is learned.

More generally, we can view all data that is represented in the network with a memory weight that is greater than zero as having been learned. This seems reasonable: as we shall see, the greater the weight, the easier it is to "find" the node in the network (and so recall the information associated with it).

As in our previous work, we assume that data strings can be compared for similarity (e.g., in the string ABMNFABXY, the second AB is recognized as being the same string, or at least a string very similar to the first AB). This is a significant assumption, because the same input may be received from different sensory nodes and comes at different sizes or intensities. Indeed, the process of what determine similarity is by itself a complex issue. However, we feel that this assumption is justified, because similar sensory inputs should eventually lead to similar profiles of activations of nodes and edges in some area of the network. (We discuss this issue at greater length in Section 4.1, when we discuss how similarity is represented.)

The effect of a subsequence of the input data either matching some data already represented in the network or appearing more than once in the input sequence is that the input data string is effectively segmented into (a) the nodes representing the (data items in the) significant subsequences (the ones that matched sequences in memory or occurred repeatedly), which have higher weight and a higher activation level, and (b) the nodes representing the remaining parts of the data string, which are also represented in the network, but have lower weight, and may be linked by (a sequence of) edges to the nodes representing the subsequences of greater weight. Thus, in our model, data acquisition already involves some level of data segmentation, simply by causing some nodes and edges to increase in weight relative to others.

But this is only the beginning of the process. We assume that when the nodes representing a subsequence are simultaneously activated, there is likely to be a particular node n in the network that will receive signals from all of them (either through edges connecting them directly or after passing through several other nodes and edges).[5] This assumption is in line with previous ideas in neurocomputation, such as the suggestion that sequential firing of nodes in the brain can be used to represent compositionality (Abeles, Hayon and Lehmann, 2004). Because the activation of this node indicates the co-activation of this group of nodes, it would also effectively represent the occurrence (i.e., activation) of this subsequence of the data input. We can use this process to construct hierarchies, where a single node represents a group of nodes, each of which represents a group of nodes, and so on. The construction of such hierarchical representations in our model captures various processes that have appeared in the literature under the names *configural learning* (Pearce, 2002; Duncan *et al.*, 2018), *chunking* (Gobet *et al.*, 2001), or *segmentation* (Brent, 1999), all of which are quite similar, and involve the learning of configurations, patterns, or hierarchical structure in time and space (Goldstein *et al.*, 2010). For convenience, and in accordance with previous work (Gobet *et al.*, 2001; Jones, 2012; Kolodny, Edelman, and Lotem, 2015a), we use the term *chunking* to describe this process, and the term *chunk* to describe the group of nodes represented by a single node, which we call a *chunk node.*

---

[5] Obviously, in a real network, there will not be a unique node available to represent each possible combination of nodes; the number of possible combinations is far too high. But it does seem reasonable to assume that there will be sufficiently many nodes to be able to represent groups of nodes that frequently fire (almost) simultaneously. An available node is used to represent whichever combination of nodes fires first.

Once a chunk node has sufficiently high weight, making it unlikely to decay, the data sequence that it represents is also unlikely to be segmented any further. This is because the repeated activation of the chunk node to reach fixation makes it likely that the weight of the nodes and edges leading to it from the nodes of the chunk also increased in weight, and thus they are also likely to have reached fixation. Thus, if a partial match to a chunk appears in the input, for example AB is observed when there is already an ABC chunk in the network, it may result in activating an AB node (a "meeting point" of A and B that is different from the ABC chunk node) but it cannot change the already fixated structure of the ABC node and the edges leading to it.

As suggested earlier, we view nodes that have high weight as representing meaningful units; we would thus want them to be retained in memory. Whether a chunk is represented by a node and retained in memory depends on the parameters of weight increase and decrease (which themselves may depend on the current activation levels) and on the fixation threshold. These parameters determine a window for learning, namely, the times between which the weight of the nodes and edges representing a subsequence and the chunk node activated by them are so low that the chunk effectively disappears, and the time that these weights are sufficiently high so that the chunk is fixated. During this window, data strings can be segmented as a result of differing weight increases of portions of the string and the formation of chunk nodes, and the chunk nodes may either increase in weight and get fixated in memory or decay and thus be forgotten.

The learning process described above is quite sensitive to the distribution of input data. Consider a word such as "backpack", which would be fixed in memory and represented by a single chunk node if it is heard repeatedly with no prior exposure to instances of "back" or "pack" (not even within other sequences, such as "on my back" or "in the pack"). On the other hand, if "back" or "pack" are heard often (before the "backpack" node get fixated), then their partial commonality with "backpack" would result in "backpack" being segmented into "back" and "pack", each represented by a different chunk node with a directed link between them (i.e., back → pack). Whether the fixation of long sequences is good for a learner is domain dependent (see (Lotem and Halpern, 2008; Kolodny, Edelman,and Lotem, 2015b)). In any case, note that the fixation of "backpack" does not prevent the formation of separate nodes for "back" and "pack" following later observations.

In addition to breaking up data strings into smaller segments represented by chunk nodes, larger chunk nodes can be created by the concatenation of smaller chunks that are repeatedly observed in succession. For example, if chunk nodes representing ABC and DEF are repeatedly observed in succession, and thus repeatedly co-activated (one soon after the other), there is likely to be a node in the network receiving signals from both chunk nodes, so that after repeated activation and weight gain it will be fixated as the "ABCDEF" node. Going back to our "backpack" example above, it may be the case that there is a chunk node representing "backpack" even if "back" and "pack" are already fixated. Similarly, a common expression such as "have a nice day" may be represented as a chunk node that is a concatenation of other chunk nodes. This process can build hierarchies of chunks in the network, a point to which we return later. Thus, chunks can be formed directly from the raw data input, or by the concatenation of previously learned chunks, creating a hierarchical structure.

The parameters of weight increase and decrease, which determine how likely parts of strings are to be represented by a node in the network, can be viewed as determining a test of statistical significance: with the right choice of parameters, natural and meaningful patterns that recur reasonably often are likely to pass the test, while spurious patterns decay and are likely to be forgotten.

A simplified example of the process of chunking and network construction, showing the critical role played by the parameters of weight increase and decrease and by the distribution of data input, is discussed and illustrated in the following section.

**3.3.3** A simplified example

To better understand the process of chunking and network construction, we consider a simple example (based on one in (Lotem *et al*., 2017)). To make this simplified example easier to follow, we assume that the transducers (the nodes and edges) are deterministic rather than probabilistic. The logic of the process is simple. It should be easy to verify the outcomes that we claim by carefully following the rules specified in the text and in the figure caption; there is no simulation or a hidden process that produces the outcomes. Figure 1A shows the network that is constructed as the result of getting three specific strings of data as input, under the assumption that the weight-increase parameter is 0.4, the fixation threshold is 1.0 (which means that a data item reaches fixation after three successive observations, since $3 \times 0.4 > 1$), and the weight-decrease (decay) parameter is 0.005 (i.e., the weight of a data item that is not yet fixated decreases by 0.005 after each time step that it is not observed, where we identify a step with a symbol being observed). We also assume that the three data strings in this illustration are separated by 30 additional (irrelevant) characters (so that 30 additional but irrelevant symbols, such as xyxw…., are observed after each of the first and the second strings).

The figure demonstrates that repeated sequences within each string (highlighted with different shades of color for clarity) are chunked due to their similarity. (The similarity leads to repeated activation and a resulting increase in weight relative to the remainder of the string.) Directed edges represent past associations between the strings represented by these nodes; the weights of these edges represent statistical regularities of the environment. For example, 98 always follows 756 and precedes 136, while 756 leads to 48361, 98, and 28 with equal probability. Despite the simplicity of this network, we can already observe that 98 and 28 have a similar link structure: both are preceded by 756 and followed by 136. In our earlier work (Kolodny, Edelman, and Lotem, 2015a; Kolodny, Lotem, and Edelman, 2015), we showed how such similarity in link structure can be used for generalization, for the construction of hierarchical representations, and for creativity (Kolodny, Edelman and Lotem, 2015b).

We stressed earlier that, according to our model, the coordination between the learning and data-acquisition mechanisms and their evolution in response to typical input characteristics are critical for building an effective network. This point is illustrated by Figure 1B, where the same data as in Figure 1A are now distributed differently, leading to a radically different network representation (although the weight increase and decrease parameters remain the same). The distribution of the data input in Figure 1B leads to the fixation of large idiosyncratic chunks of data, because each string includes three identical subsequences of eight characters that occur repeatedly within a string and hence within a short time frame (specifically, AAA, BBB, and CCC; see the figure and captions). One effect of having these large idiosyncratic chunks of data is poor link structure, which may hamper further learning and generalization (Lotem and Halpern, 2008; Kolodny, Edelman, and Lotem, 2015b). For example, no generalization can now be drawn for 98 and 28, because each of them is "locked" within another chunk. Recognizing potentially meaningful patterns (such as 98 and 28) in novel input becomes more difficult (i.e., is less likely) if the memory representation is based on large idiosyncratic chunks. The learner is then less likely to place novel data in context and to perform further useful chunking (see further discussion in Section 4).

If we change the weight-increase parameter from 0.4 to 0.3, then the data input of Figure 1B would result in exactly the same network as in Figure 1A; all the problems that we have just described would disappear. This is because the segment "75648361" would not reach fixation after the first data string and would not decay completely before the second string is acquired, so it would be segmented when 756 is encountered in the second string and again in the third. This example shows how different combinations of data distribution and weight increase and decrease parameters can affect the network that is generated. It also demonstrates how relatively simple modifications to the weight parameters or the distribution of data input can lead to major changes in the network.

In this simplified example, the weight increase and decrease parameters were fixed in each case. This might happen in a situation in which evolution has set the parameters in relation to the expected data distribution to obtain the best network. In other situations, the weight increase and decrease parameters may be adjusted over a much smaller time scale (e.g., when an agent is under stress) so that data strings that are important but rare can also be chunked. This is possible in our model because the weight increase and decrease parameters are also affected by the current levels of activation, which in turn are modulated by a range of sensory signals, indicating various physiological, social, and emotional states (LaBar and Cabeza, 2006; Lotem and Halpern, 2012).

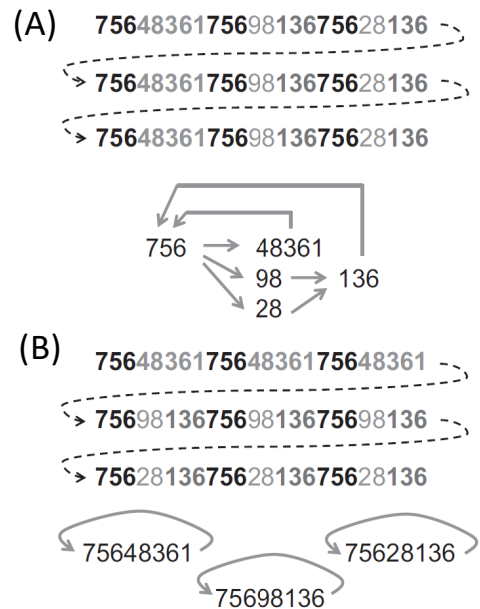

**Fig. 1.** Data input in the form of three strings, and the network that is constructed as a result of acquiring and processing this input using the learning mechanisms and parameter set described in the text. (A) Each data string of 24 characters is composed of three non-identical subsequences of eight characters that share some common segments (highlighted using the same shade of gray). The three strings are identical in this case, so labeling each subsequence of eight characters as A, B, and C, respectively, allows describing the structure of the input as ABC ABC ABC. (B) The same input as in A is distributed differently over time, which can be described in short as AAA BBB CCC. This input leads to a completely different network structure due to fixation of A, B, and C as long eight-character chunks. The weights of the nodes and the links of the networks are not shown.

## 4. How the network represents knowledge

In the previous section we discussed how the brain could construct a basic network. Here we consider more sophisticated operations that will be needed for the network to represent knowledge well. This includes issues like representing and detecting similarities, representing hierarchies, and dealing with multiple representations of the same data item.

**4.1 The representation (and detection) of similarities**

Being able to detect the fact that a portion of the input is similar to data items already represented in the network plays a significant role in our approach. For example, the fact that the sequence BC in the input ABCD can be matched with a pre-existing representation of BC allows us to effectively segment the input into the sequences A, BC, and D by
27In this simplified example, the weight increase and decrease parameters were fixed in each case. This might happen in a situation in which evolution has set the parameters in relation to the expected data distribution to obtain the best network. In other situations, the weight increase and decrease parameters may be adjusted over a much smaller time scale (e.g., when an agent is under stress) so that data strings that are important but rare can also be chunked. This is possible in our model because the weight increase and decrease parameters are also affected by the current levels of activation, which in turn are modulated by a range of sensory signals, indicating various physiological, social, and emotional states (LaBar and Cabeza, 2006; Lotem and Halpern, 2012).

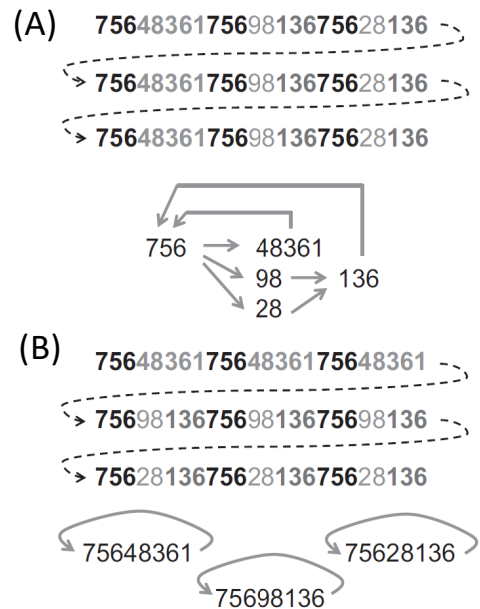

**Fig. 1.** Data input in the form of three strings, and the network that is constructed as a result of acquiring and processing this input using the learning mechanisms and parameter set described in the text. (A) Each data string of 24 characters is composed of three non-identical subsequences of eight characters that share some common segments (highlighted using the same shade of gray). The three strings are identical in this case, so labeling each subsequence of eight characters as A, B, and C, respectively, allows describing the structure of the input as ABC ABC ABC. (B) The same input as in A is distributed differently over time, which can be described in short as AAA BBB CCC. This input leads to a completely different network structure due to fixation of A, B, and C as long eight-character chunks. The weights of the nodes and the links of the networks are not shown.

## 4. How the network represents knowledge

In the previous section we discussed how the brain could construct a basic network. Here we consider more sophisticated operations that will be needed for the network to represent knowledge well. This includes issues like representing and detecting similarities, representing hierarchies, and dealing with multiple representations of the same data item.

**4.1 The representation (and detection) of similarities**

Being able to detect the fact that a portion of the input is similar to data items already represented in the network plays a significant role in our approach. For example, the fact that the sequence BC in the input ABCD can be matched with a pre-existing representation of BC allows us to effectively segment the input into the sequences A, BC, and D by increasing the weight of BC above that of A and D. Here we go into more detail about how similarity detection can be done in our framework. What makes this nontrivial is that, in our framework, there no algorithm that runs on top of the network that can search the network or compare sequences; this must be done using only the algorithms that run at each node and edge.

Consider first an idealized situation where two completely identical inputs are presented to a sensory system under identical conditions, so that they elicit identical *activation profiles;* that is, the activation level of each node and edge is the same for both stimuli. The fact that the same nodes and edges becomes more active when the second input is presented as when the first input is presented can be viewed as the second input being "matched" with the first input, although no search and comparison was performed.

The matching process works essentially the same way for inputs that are similar but not identical. Clearly, in the real world, two inputs are never identical. Similar inputs will typically not activate exactly the same sensory nodes (as most sensory systems have many sensory cells). Yet even if the nodes and edges that are first activated by similar inputs are different, if the inputs are similar, we expect that somewhere along the activation cascade the same nodes and edges will be activated. If no such shared activation occurs, then there is nothing in the process that indicates that the inputs are indeed similar. On the other hand, if repeated observations of similar inputs result in the repeated co-activation of the same group of nodes and edges, then this group of nodes and edges can be viewed as representing the similarity of the two inputs. Not only will the repeated activation of these nodes and edges increase their weight, but it is also likely to increase the weight of the node receiving signals from all (or most) of them, which will now become a "chunk node" representing the similar inputs. For example, if the inputs A and A' are not identical but only similar, the agent will recognize them as similar if there is a chunk node that fires in response to receiving either input (which means that it receives sufficiently strong signals from a sufficiently large number of nodes in both cases). When we say that the agent recognizes them as similar, we mean that if A is associated with certain data and elicits certain responses, A' will have the same effect because both inputs activate the same chunk node in the network.

Now consider a new input of a data string that contains the sequence A". If A" is sufficiently similar to A and A', it will activate many of the same nodes and edges as A and A', so the chunk node representing A and A' will fire. This is how "matching" is done. The detection of similarities (and the matching of similar inputs) in our framework amounts to the activation of shared nodes and edges.[6]

---

[6] The logic of this principle is clearly not new; it has been applied to detecting similarities by neural networks (Rumelhart, Hinton, and Williams, 1986).

The probability of finding a match (i.e., finding similarities) depends on several factors. First, it increases with the extent of activation, since lower activation results in fewer nodes being reached (i.e., fewer nodes having their activation level increased), making it less likely that a shared node is activated. Second, the probability that the activation elicited by similar inputs will hit the same nodes and edges in the network is affected by the physical structure of the network, and the extent to which different paths lead to different nodes. In a "well-constructed" network (i.e., one which is a good representation of the environment), we do not want too many different paths leading to the same nodes. This would lead to too many different inputs activating the same nodes, and thus too many things being viewed as similar.

There are clearly additional important aspects that determine how the network represents similarity and which are considered to be sufficiently similar. For example, earlier, when we described the process of network construction, we implicitly assumed that similarity is sensitive to the order in which elements appear in the data. For example, ABC and ACB were viewed as different data sequences. This may be important for some data types, such as the sequence of actions of a particular hunting technique, the phrases in a birdsong, or human speech. But for some other types of data it may be sufficient (or even better) to classify two sequences as similar by merely recognizing some of their similar components. For example, two instances of the same salad in a salad bar or a stand of mango trees in the forest may be recognized based on a combination of stimuli, ignoring their exact serial order (which may vary across instances). Clearly, when the exact serial order is not important, the similarity criterion is broader. How can the process of matching as we have defined it possibly distinguish between the two cases?

There are several solutions to this problem of "serial order coding". We do not claim to know how exactly it is done in the brain (and it may be done in more than one way; see, e.g., Ghirlanda *et al*., 2017, for a recent review). A plausible assumption is that in most networks, there are nodes that differ in how the serial order of stimuli affects their activation level. Consider for example the input {A, B}, which may be observed as "AB", as "BA", or as a simultaneous appearance of the two. If there is a node n in the network receiving signals from both A and B (i.e., has incoming edges from both), its level of activation will increase most if the two signals arrive at the same time. If one signal arrives a bit later, the increase is likely to be smaller because the effect of the first signal has already started to decay. However, if a node n' in the network is positioned in a way that causes the signal from A to take a bit longer to arrive (e.g., the signal coming from A goes through an intermediate node before reaching n'), then its level of activation will increase most if A is observed somewhat before B, because then the two signals will arrive together. In other words, the activation level of node n' increases most in response to "AB". Similarly, there might be a node n" for which the signal from B takes a bit longer to arrive, causing its activation level to increase most in response to the input "BA". Since an increase in activation level leads to an increase in weight, nodes n' and n" will represent the occurrence of "AB" and "BA", respectively, while the node n for which the signals from A and B arrive at the same time represents the simultaneous occurrence of A and B.

The node n is particularly likely to represent the simultaneous co-occurrence of A and B if the increase in activation level at n decays rapidly, so there is unlikely to be much of an increase in weight if the observations are not simultaneous. A slower decay may allow the formation of a chunk node that is not sensitive to the serial order of A and B and may therefore represent any co-occurrence of A and B within a relatively short time interval. If AB and BA are observed with roughly the same frequency and in random order, this order-insensitive node will gain weight and be the first to reach fixation. With this representation, an input of any configuration of AB would result in an activation of this order-insensitive {A, B} node. This is not to say that order-sensitive chunk nodes representing AB and BA cannot be created as well. If "AB", for example, is observed repeatedly, then the node n', which is sensitive to order, may also get fixated. Indeed, if "AB" is observed far more often than "BA", it is possible that n' will be fixated and n will not be. In any case, note that the input "AB" may well activate both the {A, B} and the AB chunk-nodes, implying that it can be recognized in several ways.

The fact that several nodes can recognize the same input is important beyond the issue of serial order. This is how the fact that an observation is an instance of a larger set (or, more generally, in instance of several nodes in a hierarchy; see Section 4.2) can be recognized by the network: A specific dog may be recognized both as "Ginger" and as a dog, and the text of the word "banana" can be matched with a real banana, as well as with "word in the English language", indicating that the text is in English rather than in Japanese.

The process of finding similarities in incoming data streams is made even more complicated by the fact that some sequences are likely to include several overlapping similarities (and hence alternative matches). For example, the input ABCDEF may be similar to the pre-existing chunks BCD, BC, and DEF. How does the matching process decide which is the best match? All these pre-existing chunk nodes are likely to be activated, although only one of them may be the "correct" match given the current context. For example, hearing "carpenter" may activate the pre-existing chunks "car" and "pen", hearing "understand" may activate "under" and "stand", and so on.

In general, when there are multiple matches, so that many nodes are activated, we expect the network to have a way to decide what the "right" matches are; for example, we don't want the weight to increase for all the matches, but only for the "right" ones. Although decision-making in general will be addressed in Section 5, some form of decision-making is already necessary to handle this problem, so we briefly consider it here. Suppose that an agent observes a data string containing multiple, partially overlapping matches. It is reasonable to expect that if several chunk nodes are activated by the input, then the one representing a chunk that is most similar to one of the subsequences in the input will have the greatest activation level. For example, if the input is "**A**BCD**EF**", BCD is likely to have a higher activation level than D**EF** (which is similar but not identical to the pre-existing chunk DEF, because the **E** and **F** are thicker). Therefore, the increase in the activation level, and consequently in the weight of BCD, would be higher than that of DEF. But second, and more important, even if an input activates several nodes in the network, the effect on weight increase (and hence on memory and learning) also depends on the initial activation level of these nodes. That is, if the node representing the chunk BCD was currently very active but the nodes representing BC and DEF were not, then, the process of data segmentation and acquisition would take place mostly around the BCD chunk (and the effect on the portions of the network that include the BC and the DEF chunks may be negligible). Using levels of activation for choosing among alternative matches is adaptive. The differences in current activation represent the current state; what the network has been currently responding to and what is expected to come next (see our discussion of prediction and prediction error in Section 5.1). The level of activation can therefore help predicting which matches are more likely to be relevant in the current context, and which should be ignored.

In Section 5, we consider more refined versions of decision-making processes. Both the version discussed above and the more refined version discussed in Section 5 become more complicated in this case. Note that, statistically speaking, multiple matches are more likely to be found if the activation of nodes and edges by a new observation affects a relatively large number of nodes and edges. Thus, setting the activation-level parameters to allow rapid increase in activation levels can have the downside of increasing the problem of multiple matches. Moreover, recall that in our framework, the working memory used to process incoming data is defined by the number of nodes and edges activated by the new data. This implies that although increasing the size and the duration of working memory (by allowing the activation levels to affect more nodes and edges for a longer time) increases the chances of finding a match, it can also make the matching process more challenging, since more matches will be found, including ones that are unlikely to be relevant to the context. In earlier work (Lotem *et al.*, 2017), we suggested that the computational load of checking all possible matches in incoming data is especially high when serial order is important, as in the case of language, which may explain the evolution of limited working memory in humans. So far we have not explained why checking and deciding among all possible matches is costly and therefore favors a limited working memory; we consider this issue more carefully towards the end of Section 5.3 when discussing decision making.

**4.2 Context, categories, and multiple representations**

In our framework, the process of matching incoming data with pre-existing nodes in the network can be viewed as a process of putting new data in context, which can be viewed as "understanding" the data. In this subsection, we consider how this can be done.

We take the *context* of a data item that is represented by a node in the network to be the set of edges associating this node with other nodes in the network. These associations (and their relative strength, as determined by the weights of the nodes and edges involved) represent all the information that has been learned about this data item, and thus can be viewed as its meaning. For example, a node representing "apple" may have links to "apple tree", to the act of "eating", and to the color "red", which means that an apple is understood (based on its context) as something that is found on a tree, can be eaten, and may be red. When a new "apple" is observed, matching the new "apple" with the pre-existing node of "apple" in the network puts the new "apple" in the context of previously observed apples, and thus can be viewed as giving the observation its meaning.

The context of a node often has a natural structure. For example, as we observed earlier, the repeated co-activation of chunk nodes can create a hierarchical structure (whose elements consist of chunk nodes); this is how the network can capture the fact that a tree is composed of leaves, branches, and a trunk; a forest is composed of many trees; a landscape is composed of a forest and a lake; and so on. Note that such a hierarchy reflects the agent's understanding that a branch can be understood as part of a tree, which can be understood as part of a forest.

Another type of hierarchical structure that emerges quite naturally is what has been called an *is-a hierarchy* (Brachman, 1983). The first step in creating an is-a hierarchy is for clusters that have a similar link structure to be represented by a chunk node. We can think of the chunk node as representing a *category*. For example, Danny, David, and Jennifer may all share features and attributes common to children; in addition to individual nodes representing each of Danny, David, and Jennifer, there may be a chunk node to which they are all linked that represents "children". This can occur if the nodes representing individual children are activated almost simultaneously sufficiently often; for example, the nodes representing many of the children may be activated simultaneously at a soccer game, or if there is some other activity involving these children. Similarly, an orange, banana and mango may all share the common attributes of fruits, and thus have a similarly link structure. As a result, they will all become more active if one of their common attributes is observed. If this happens often enough, a chunk node will be created that we can think of as representing "fruit". Note that while the data items themselves may not be similar (e.g., bananas and oranges are quite different in appearance), the similar link structure of the nodes that represent them gives them similar context or meaning (e.g., edible sweet object that grows on trees), and results in the creation of a chunk node. (Although in human language such categories may often have a name, such as "children" or "fruit", this may not always be the case, and may not be necessary for recognizing the data items of such clusters or categories as being similar in some sense. For example, the similar link structure of different nodes representing different words in spoken language allows using them as synonyms (Kolodny, Lotem and Edelman, 2015) without having an explicit word describing their category. Similarly, animals can generalize across many different prey items without ever learning the word "prey" or "food".)

In any case, once we have chunk nodes representing categories, chunk nodes with a similar structure can themselves be represented in a hierarchical fashion by chunk nodes. The upshot is an is-a hierarchy: A mango is-a type of fruit, fruit is-a type of edible object, and so on. Such is-a hierarchies can be used for making generalizations. Indeed, previous implementations of our framework showed how similar link structure can be used for making generalizations, planning, and problem solving (Kolodny, Edelman, and Lotem, 2015a, 2015b; Kolodny, Lotem, and Edelman, 2015). The critical step in making such generalizations possible is the assumption that if the link structure of A and B is "sufficiently" similar (e.g., both have edges leading to E, F, and G), then a known feature of A (e.g., an edge leading from A to X) is also attributed to B (e.g., an edge leading from B to X is added to the network). The logic of this step seems intuitive; the features of a member of a category are generalized to all members of this category. However, to explain exactly how this can be done just using of the programs run by the nodes and edges of the network, we need to discuss the process of planning and decision-making. We thus defer further discussion of generalization to Section 5, where we discuss planning and decision-making.

The process of network construction may result in multiple representations of the same type of data input. Some multiple representations may arise because similar subsequences in data input arrive in very different settings, so are not recognized as being similar by the matching process. Thus, they will be represented by different (clusters of) nodes in the network. For example, at some point a child might have several different nodes representing dogs. Some of these may represent the same dog as a result of viewing the dog from different angles or viewing the dog at different times without realizing it is the same dog. But after additional observations, these nodes will typically coalesce into one node representing the same dog. More precisely, certain nodes that represent the same dog will all be activated at essentially the same time, so a chunk node will be created that represents these nodes. Many or all of these nodes (i.e., the original nodes and the chunk node) may even become fixated, and some may decay, but since all the ones that persist will fire simultaneously and will be connected to the same nodes, they can be understood as representing the same object. On the other hand, nodes representing different dogs are likely to have somewhat different link structures (the dogs may have different colors, or different sizes and shapes, or have different owners). Thus, these nodes will not always be activated at the same time. There may be a chunk node representing the category "dog" to which all of these nodes are linked, although the network structure will still reflect the fact that they are different dogs.   We can also have a more general an is-a hierarchy involving dogs (Rover is-a terrier, a terrier is-a dog, a dog is-a mammal, …).

**4.3 Multiple representations and episodic memory**

Multiple representations of data may be especially important for representing different types of statistical relationships. The process described so far (of segmenting incoming data and updating the weights of nodes and edges in the network) creates a statistical representation of the environment, but does not preserve *episodic* memories (i.e., the exact historical sequence of events). The statistical representation that aggregates observations over time is part of what has been called *episodic* memory, as distinct from *semantic* memory (Tulving, 1972, 2002). Episodic memory and semantic memory serve different needs. Separating between them was probably necessary as soon as foraging animals had to keep track of particular recent events without confusing them with an aggregated representation of similar past events (Arbilly and Lotem, 2017). For example, for a bird that feeds on grasshoppers, it may be critical to distinguish between the general likelihood of finding a grasshopper under a rock (which may be low in general) and the probability of finding it under a rock a minute after it was seen scurrying near the rock. In terms of our model, separating semantic from episodic memories implies that there must be nodes representing the semantic memory and nodes representing the episodic memory.  We expect that the nodes representing semantic memory will be chunk nodes, created by the process of matching and segmentation, as described above.  These chunk nodes represent statistical information gleaned from repeated observations. They do not necessarily preserve episodic information, because the links to previous and subsequent nodes may be aggregated; for example, we cannot tell whether a chunk representing the sequence A→B→C→D was formed by an episode such as A→B→C→D or by a combination of the episodes A→B→C and C→D.  On the other hand, a sequence such as A→B→C→D might well represent episodic memory.  In general, like semantic memory, we expect episodic memory to be represented by chunk nodes; observations of the same object at different times will thus be linked to different chunk nodes.  For example, if the same type of cereal was eaten for breakfast on Tuesday and on Wednesday, there would be a link from a node representing the cereal to a chunk node that represents the episodic memory of breakfast on Tuesday, and another link to a chunk node that represents the episodic memory of breakfast on Wednesday. Although in our framework we do not have special "semantic-memory nodes" and "episodic-memory nodes", they can arise naturally from the programs run by nodes and edges, just all other chunk nodes do. The problem would then be to distinguish between them, which can be done by their profile of edges. Episodic memories are linked to memories from the same time frame, which may not be similar in any way.

Note that nodes that represent episodic memories are likely to decay rapidly. Each sequence of events is typically idiosyncratic and unique, so that repeated activation of the same episodic node by further sensory input is unlikely. That said, there are two ways in which episodic nodes might become fixated. First, some episodic nodes may be co-activated with a reward node as a result of an especially fearful or rewarding context. (Recall that we assumed that such reward nodes were present in the initial network). This may result in an activation that is strong enough to reach memory fixation after only one observation. Second, a related possible mechanism that leads to the same result is based on the idea that memories may increase in weight as a result of *self-reactivation*: a process, also called *replay*, that can result in memory consolidation (reviewed by (Carr, Jadhav and Frank, 2011)). In terms of our model, this process implies that agents reactivate the nodes and edges involved in an episodic representation, which increases their weight until they reach memory fixation. In our framework, we expect reactivation to occur following important events, such as soon after experiencing a strongly reinforcing event, either positive or negative, or when exposed to a novel environment (Carr, Jadhav and Frank, 2011). If there is a cycle among the edges (e.g., from A to B to C back to A again), we will have such reactivation without further sensory input. This is especially likely if the event had an initial high level of activation and the agent keeps thinking about it over and over. This reactivation process will greatly increase the level of activation of the relevant nodes and edges (as well as increasing their weight), which means that stimuli (either external or internal) that normally would not cause that nodes and edges to be reactivated will do so. We are all familiar with this phenomenon: everything reminds you of food if you are hungry; everything reminds a broken-hearted lover of his/her former partner after a breakup; driving home after a job interview, free association always leads you back to thinking about the interview. There is, of course, the concern that this process will "over-react", and result in too many episodic memories being retained. That is why it is important that the parameters are set appropriately by natural selection.

We would expect an interaction between nodes representing episodic and semantic memory. Such an interaction might be important for updating the network: reactivating related episodic memories may be used to aggregate them in order to create a new semantic memory. For example, after several commutes to a new working place, the repeated commute episodes (or, at least, the parts of them that have no edges to unusual events) may be forgotten and be replaced by a sequence of nodes and edges representing the whole commute.

# 5. Computing and acting: from knowledge representation to behavior

So far, we have outlined how the probabilistic transducers can construct a network that represents learned information. Here we explain how the same transducers, and the network they form to represent knowledge, can also compute and act using this network representation to produce the type of behaviors that would be expected from the brain of an autonomous organism. As we shall see, activation levels play a critical role.

We start with prediction and prediction error. We then consider how the network as we have described it can perform simple actions and make decisions when it receives an input. We then work our way to more complicated tasks, such as planning and executing plans. We conclude by trying to address some of the more difficult questions, such as those related to setting goals and priorities, self-initiation and free will, and issues of consciousness and causality.

**5.1 Predictions and prediction errors**

It is has been suggested that the most fundamental use of learning is to generate predictions and to minimize *prediction error* (e.g., (Rescorla and Wagner, 1972; Friston, 2010; Hohwy, 2013)). In our framework as well, being able to use the network to generate behavior depends on the network's ability to predict. Prediction clearly plays a critical role in planning and decision-making. The following simple example illustrates one way that our networks can do prediction, and shows that they have an analogue of prediction error.

Suppose that an event E typically precedes an event E'. To be specific, suppose that E is the sound of a bell and E' is receiving food. One key assumption (that was also made in some of our previous examples) is that events are typically not represented by a single node, but by a cluster of nodes. We can think of each node in a cluster as representing a different aspect of the event. Moreover, some aspects may not always occur when the event occur. There may be, for example, a node that corresponds to the experience of actually receiving food, another node that corresponds to the experience of predicting food (as a result of hearing a bell), another that corresponds to the experience of not receiving food after expecting it, and so on.

Suppose that there are nodes Bell, Food, No-Food, Pred-Food, and Pred-No-Food, which represent actually hearing a bell, actually receiving food, not receiving food, predicting that food will be received soon, and predicting that food will not be received. How do these nodes arise? It is easy to see how sensory input can activate, and thus increase the weight of the nodes representing hearing a bell and actually receiving food. But how do the other nodes arise? Figure 2 may help the reader follow the explanation.

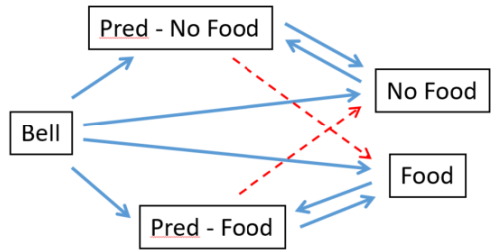

Figure 2

If hearing a bell is frequently followed by receiving food, then there will also be a node (which we can think of as the Pred-Food node) that has edges coming to it from both Bell and Food; its weight increases whenever Bell and Food are activated sequentially. (Recall our discussion in Section 4.1 of how we might have a node that detects the sequential observation of A then B.) The No-Food node, on the other hand, initially has edges coming from Pred-Food and Bell; it will be activated and increase in weight whenever Pred-Food and Bell fire but food does not appear (so Food is not activated). Finally, if hearing a bell is occasionally followed by not receiving food, then there will also be a node that we can think of as the Pred-No Food node, that has edges coming to it from both the Bell and the No Food nodes (see Figure 2). With this architecture, the weight of the Pred-Food node, and of the edge leading to it from Bell, represents how often Bell was followed by Food, while the weight of the Pred-No Food node and of the edge leading to it from Bell represents how often Bell was not followed by food. Similarly, the weight of the edges leading from Pred-Food and Pred-No Food to Food and to No Food, respectively, represent how often the predicted outcomes occurred, and the weight of the edges leading from Pred-Food to No Food and from Pred-No Food to Food (marked with red dashed arrows) represent disappointment and surprise, respectively.

We can think of the prediction error in our model as the difference between the relative activation levels of the two prediction nodes (Pred-Food and Pred-No Food) and the relative activation levels of the nodes representing the actual events (i.e., of Food versus No Food). More precisely, taking $act_t(n)$ to denote the activation level of a node n at time t, taking $t_1$ to be just after the bell rings, and $t_2$ to be the time by which food is expected, the prediction error can be written as

$$\text{Prediction error} = \text{act}_{t_1}(\text{Pred-Food}) / (\text{act}_{t_1}(\text{Pred-Food}) + \text{act}_{t_1}(\text{Pred-No Food})) -$$
$$\text{act}_{t_2}(\text{Food}) / (\text{act}_{t_2}(\text{Food}) + \text{act}_{t_2}(\text{No Food}))$$

We expect that the activation level of Pred-Food and Pred-no Food at time $t_1$ to be proportional to their weights (which, in turn, are proportional to their relative frequency of occurrence). That is, when the bell is heard, Pred-Food and to Pred-no Food are activated roughly proportional to how often they have correctly predicted food and no food in the past. Thus, the first fraction,

$$\text{act}_{t_1}(\text{Pred-Food}) / (\text{act}_{t_1}(\text{Pred-Food}) + \text{act}_{t_1}(\text{Pred-No Food}))$$

represents the relative likelihood of getting food after the bell rings, based on past experience. On the other hand, at time $t_2$, either Food has arrived, in which case the Food node is quite active and the No Food node is not active at all, so that the second fraction is 1, or Food has not arrive, in which case the second fraction is 0. The difference between the fractions is a measure of how the prediction compares to reality.

To see how this prediction error is related to the learning process in this example, we must show that smaller prediction errors are correlated with less learning (and vice versa), and that learning minimizes prediction errors. Recall that learning in our model is effectively expressed by changes in the relative weights of nodes and edges in the network. Consider first a case where food is almost always received after hearing the bell. In this case the weight of the Food and Pred-Food nodes and the edges leading to them from Bell are already high; consequently, the relative activation level of the Pred-Food node after hearing a Bell (the first fraction above) is close to 1. If food is received as expected, the prediction error is small, which means that the activation level of Pred-Food and the edge leading from it to food cannot increase much further (it was already close to 1). As a result, very little learning (change in weight) would follow. On the other hand, if food is not received, the prediction error would be large. Then the Pred-No food would be activated, leading to a significant weight increase of the No Food and Pred-No Food nodes, and the edges leading to them, and also affecting the predictions for the next event. (The extent of the effect depends on the parameters of weight increase and decrease.) Importantly, if food is not received next time either, the prediction error would be smaller, since the first fraction, which roughly corresponds to the relative frequency of food to no food after the bell, would now be smaller than after the first time that food was not received. In that sense, we can view learning in our model as minimizing prediction errors.

A similar process occurs when food is expected with some intermediate probability after ringing the bell. The expectation moves in the direction of what actually happened, thus minimizing prediction error if the recent change in the probability of receiving food persists.

Obviously, the details of the process depend on many assumptions we can make regarding how exactly activation levels are affected by relative differences in weights and on how weight increases with further observations. (Our discussion in Section 2.4 of how weights increase after they go beyond the threshold is quite relevant to this issue.)

Although it has been suggested that, from a normative point of view, the minimization of prediction error is the goal of learning (Friston, 2010), in our model this minimization is the outcome of the rules of the programs of nodes and edges that we expect to evolve to construct a useful representation (i.e., a representation that can support behaviors that increase biological fitness).

**5.2 Simple actions**

When we first described how probabilistic transducers can form a network, we argued that the networks developed from previously evolved stimulus-response mechanisms (Section 3.1). A stimulus-response mechanism does more than just represent information; it can act (at least, at a primitive level) in response to stimuli. More generally, we view our network not just as a representation of information, but as a program that can respond to stimuli based on the information encoded in the network. Since the network is connected to external effectors, it can also execute actions. In other words, an input leads to activity in the network that may eventually result in behavioral output. The details of the process of going from input to output depend on the structure of the network (and thus on information learned from past experience) and on the state of the network (which may be affected by other input, and also by previous experiences, since they affect the activation levels and the weights). While this process can lead to highly complex dynamics, it is relatively easy to see how it works in the case of simple actions and decision-making.

For example, a chick is born with an innate tendency to peck at small objects and to try to eat them. Initially, the decision regarding whether to peck and whether to eat is based on innate responses to pre-specified shapes, structures, and tastes, as well as to physiological states, such as hunger. As we said earlier, the innate responses can be captured in the initial innate network; the actions can be captured by having nodes connected to effectors. As soon as the chick gains experience, the same decisions are also affected by the range of stimuli that were associated with these innate components, and by the associations of these stimuli with other learned or innate stimuli. For example, the chick may learn very quickly that brown grains are tasty but white grains are not, and that brown grains can be found under green leaves at the base of some tree trunks, and that the sound of repeated pecking by other chicks may indicate that brown seeds were found, and so on. While this transition from innate responses to learned behaviors is quite intuitive to anyone with some knowledge of animal behavior, we have to describe explicitly how this process can be captured by our model.

Both the innate and the learned parts of the network in our model consist of nodes and edges. The difference is that the innate parts of the network already have high weight at birth, while the learned parts start with minimal weight and gain their weight only as a result of the learning process. Having high weight at birth implies that the innate nodes and edges that form the innate parts of the network offer built-in associations between nodes. Some of these associations lead from sensory nodes to other nodes, which in turn are connected to other nodes, and eventually to effector nodes. This is how the innate stimulus-response mechanism is captured in our network model. As mentioned earlier, some of the innate associations lead to reinforcement centers that provide feedback such as pleasure, pain, fear, or disgust. In terms of our network, these centers may be viewed as nothing more than some innate portions of the network that, upon receiving input, result in output. We would expect that, among other things, the edges coming out of these reinforcement centers have evolved to be associated directly or indirectly with effector nodes that elicit the relevant adaptive responses (e.g., pain should lead to rapid avoidance or escape, disgust should lead to throwing up, and so on).

Consider a process that is initiated by a sensory input from the environment. For example, the chick may observe a brown grain. We assume that such input from the sensory system comes to a particular portion of the network, and results in its activation level increasing. The profile of this enhanced activation pattern represents the input. For example, one profile might represent a brown grain, while another might represent a white grain. We also assume that there is an innate template for grain-like objects in the chick's network (i.e., a node or a collection of nodes and edges connecting them whose activation level increases when a grain-like object is observed). Observing a brown grain activates these nodes, which we can assume are associated with effector nodes that initiate the simple innate action of pecking and eating.[7]

So far, we have merely described how a simple stimulus-response process works in our framework. But it is quite easy to see how this innate portion of the network can be extended through learning. For example, as a result of co-activation, the "brown" color of the grain will be associated with the general template of grain-like objects. Following the outcomes of pecking and eating it, which were positive in this case (i.e., they are linked to "tastes good"), it will also be associated with the representation of tasty food (which is associated with a particular reinforcement center). A similar encounter with a white grain may result in associating "white" with grains, but in this case also with "bad taste". As part of the organism's stimulus-response set, bad taste (or merely the lack of a good taste) is innately connected with a reinforcement center that elicits appropriate actions; in this case, the appropriate action would be avoidance. Thus, the learned information represented by nodes and links that are not part of the initial innate network (e.g., "brown", "white", and the edges connecting them) clearly improves the innate response. Note that, in this example, it is clear how the network "knows" what to do. It initially knew what to do; now it can do it better, or more precisely, it can react to a wider range of stimuli.

---

[7] Such innate behaviors, commonly known as "instinctive behavior" or "fixed-action patterns", must also be encoded in the network, possibly as nodes that arrange for effectors to fire in a particular sequence.

This basic example helps demonstrate how a network that represents learned information can also compute and act to produce appropriate behavior. Upon seeing a white grain, the representation of "white grain" (or "white" and "grain") is activated, and signals are sent through the edges leading out from the nodes of this representation to other nodes and edges in the network. The increase in the activation level of the other nodes and edges depends on the current strength of the signal and the weight and the initial activation level of these nodes and edges, which is determined by previous experiences. The effect of these quantities on the transmission of signals creates a decision-making process that eventually determines what action will be taken. If the link between "white" and "bad taste" has relatively low weight (which will be the case for young chicks, for whom the combination was experienced only once or twice), it is quite likely that the chick would still try to eat the white grain rather than avoiding it, because "grain" is still innately (and strongly) connected with the action of eating (i.e., the weight of the "grain-eat" link is still higher than that of the "white-bad taste" link).

## 5.3 Planning and decision making

The description above assumes that an external sensory input initiates a process that ends with an action (seeing a grain and responding to it). A more complex process may be initiated by an internal sensory signal, such as a hunger signal coming from the digestive system. We would expect such a signal to lead to some search, planning, and execution of plans. Note that the terms "search" and "planning" in our framework merely mean an increase in the activation level of nodes and edges that results in some action. We do not have special algorithms for searching or planning. The difference between "search" and "planning" may also be a matter of perspective: when a hunger node activates nodes and edges that were associated with it in the past, it can be viewed as initiating a search for food. This will become clearer as we go through some examples.

Suppose that a chick gets a hunger signal. This activates related areas in the network. Some activation is transmitted through innate stimulus-response links going from hunger to foraging activities, such as walking and scanning the environment, perhaps until "brown grains" are found. Additional signals will go from nodes representing foraging activities or the experience of hunger to nodes representing various food types and environmental cues or objects that were associated with them in the past (see (Arbilly and Lotem, 2017)).

Now suppose that the chick forages for grains, but no grains are visible. The chick, prompted by a hunger stimulus, has to decide what actions should be taken in order to find grains. A simple response to hunger may be to keep walking and scanning the ground until grains can be seen (and then responding to the grains as described above). This response is not so different from the one described earlier, except that it is initiated by an internal stimulus (hunger) rather than by an external stimulus (grains). However, this response may not be the most effective one. The chick is more likely to find food if it can use what it sees in order to carry out a more targeted search and to plan actions that increase the likelihood of encountering brown grains. This can be done as the chick gains more experience; this experience is represented by various associations of data items.

First, the internal "hunger" signal results in the activation of nodes that represent past experiences associated with hunger and food, such as those representing the actions of eating and pecking, and a range of stimuli and actions learned in association with them. However, while a substantial part of a chick's network is likely to have at least some paths to nodes representing food or foraging behavior, we don't expect all of these nodes to be activated. The level of activation generated by the hunger signal depends on the strength of the hunger signal and on the weight of the nodes and edges associated with hunger, which means that, initially, only the most frequently experienced associations will experience an increase in activation level. The obvious problem is that the most frequent associations may not be relevant to the current state. For example, hunger and eating may be most strongly associated with the feeders at the hen house, which are not relevant for foraging in the nearby forest. The simplest way to solve this problem is to keep sending hunger signals with increasing strength until some less-frequent associations (i.e., ones of lower weight) that are nevertheless relevant to the current state and offer a potential path of action are finally activated. For example, the association of brown grains with green leaves that can be seen on the forest floor may elicit a search under green leaves. This process can easily be carried out using the network as described so far.

The network can also facilitate a more sophisticated process of planning that may be faster and more effective. At any given moment, a process of activation, similar to that elicited by the internal hunger signal, is also elicited by the external stimuli to which the chick is currently exposed. The chick sees many objects and hears many sounds that may activate the nodes associated with them in the network. If one of the nodes that are activated by the external stimuli is also activated (almost) simultaneously by the internal "hunger" stimulus, then there will be a path in the network characterized by enhanced activation levels connecting the recently activated hunger node with the nodes activated by the external stimuli. This path corresponds to a sequence of actions (more precisely, some of the nodes in the path are linked to effector nodes, which carry out actions) that is relevant to the current state of the environment. To understand why, recall that the activation coming from the hunger node can potentially increase the activation level of paths corresponding to sequences of actions that have yielded food in the past. They can represent features in the environment through which navigating and moving to the location of food is possible, or some specific actions taken in the past in order to find or extract the food (or a combination of both). Many of these sequences may not be relevant to where the chick is currently standing. However, the nodes activated by external stimuli can suggest which of these paths are most relevant to the current situation: by increasing the activation level of these relevant paths, it makes it more likely that the relevant actions will be executed. For example, suppose that the chick doesn't see the green leaves activated by the hunger signal, but does see gray tree trunks of a certain type that are associated in his memory with "green leaves" (that can sometimes be found at the base of such trunks). Thus, the activation of nodes and edges in different areas of the network (internal hunger signal and external visual input) leads to the activation of other nodes, until both sequences of activations reach the same node (the node associated with "green leaves" in the example above). This chain of activations results in a sequence of nodes and edges with a relatively high level of activation. If nodes on this path are associated with effectors that result in movement (e.g., "go to" A, then to B then to C) or effectors initiate specific actions (e.g., "dig", "chase", "peck"), a sufficiently high level of activation of this sequence of nodes and edges can potentially result in executing the corresponding sequence of actions needed for finding grains (we discuss shortly what we mean by a level of activation being "sufficiently high" for executing an action).

This process of creating paths with high levels of activation may suggest further plans when combined with the representation of categories discussed in Section 4. Suppose that our chick in the forest does not see gray tree trunks (which are associated in his memory with green leaves) but only white tree trunks (which were never seen before together with green leaves). Can it generalize the association of gray tree trunks with green leaves to white tree trunks? Although no edge exists in the chick's network between white tree trunks and green leaves, seeing white tree trunks can nevertheless raise the activation level of "green leaves". This is because both gray and white tree trunks are associated with the nodes representing the general category of "trees", whose activation level increases upon seeing white tree trunks, and is also associated with "gray tree trunks", which in turn are associated with "green leaves". Thus, both the external stimulus of "white tree trunks" and the internal hunger signal can activate the "green leaves" node, offering a novel path from "white tree trunks" to "grains" and "hunger".

Having a potential path of action represented in the network (by a high level of activation) does not mean that this path should be executed. Some paths may lead to better outcomes than others; and sometimes it might be better not to act at all and wait for a better opportunity. We would like the execution of (the actions on) a path to be the outcome of a process that tests and compares paths. Such a process can be viewed as a decision-making process. It can arise in our framework in at least two ways. First, as in a simple stimulus-response mechanism, actions are executed only if the level of activation of nodes on the path linked to effector nodes is above a certain threshold. The probability that the activation levels on a path are high enough to result in actions is related to the weights of the nodes and edges on the path (because high weight facilitates higher activation). That is, past experience (represented by the network) affects decision making via the weights: because they have a higher weight, we have a higher likelihood of executing actions that were rewarded under similar conditions in the past.

A second, more complex, decision-making process may arise when more than one path reaches a relatively high level of activation. The network must then choose the one that is best given the current context (or circumstances). To do so, the network must somehow "consider" the outcomes of each path. It is certainly not good enough to execute the actions associated with a path whose first link has the highest weight. For the chick in the example above, there might be a strong link between the nodes representing gray tree trunks and those representing green leaves (because many of the gray tree trunks indeed have green leaves near them), yet the probability of finding brown grains under green leaves may be small, which is represented by another strong link in the network associating green leaves with the experience of finding nothing. On the other hand, the chick might also hear from the other direction (away from the trunks) the sound of other chicks pecking at the ground. Suppose that the link between these sounds and the visual scene of other chicks eating is somewhat weaker than the link between gray tree trunks and green leaves, but that the visual scene of other chicks eating is associated strongly with finding grain (which is represented by a strong link to grain and a very weak link to "finding nothing"). In this case, following the initially stronger activation between "gray trunks" and "green leaves" would not be the best decision. Long-term planning involves considering the expected outcomes of increasingly longer paths (just as when playing chess).

The decision-making process must therefore be based on a mechanism that allows simulating and testing alternative paths before a decision is made (see also (Pezzulo and Cisek, 2016)). There is evidence that animals and humans replay behavioral sequences or simulate them in their brain before executing actions (e.g., (Skaggs and McNaughton, 1996; Eldar *et al.*, 2020)). How can such simulations be implemented in our framework using only the basic principles described thus far? There is no external agent that can somehow run tests of alternative paths and compute expectations.

To test alternative paths, a simulation process must somehow be generated through the activation process, and its outcomes should determine which path should be chosen. This can indeed be done, just using the principles described so far, if we assume that effector nodes produce actions only in response to input signals that are relatively high. In other words, if the level of activation must exceed a certain (relatively high) threshold in order to execute an action, repeated activation of a path may be necessary to reach that threshold. To understand how this simple assumption can facilitate planning through simulations, consider the following example.

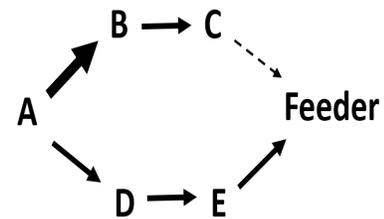

A chick standing at point A can reach a feeder via two paths. Suppose that the first path goes through B and C, and the second goes through D and E. Suppose further that the link between A and B is of high weight, but the link between C and the feeder is of low weight, so that overall, the path going through D is more likely to lead to a feeder despite the fact that the initial link from A to D has lower weight than that from A to B (see illustration). If the threshold for executing the actions "go to B" rather than "go to D" is relatively low, the chick is likely to choose the path A-B-C rather than A-D-E. This is because the initial hunger signal that activated the two paths from the chick to the feeder results in a higher activation level for B than D, so B is more likely to output a signal that is strong enough to execute the action "go to B". However, if a higher level of activation is needed for an action to be executed, more hunger signals will have to be used and to pass between nodes before an action is executed. Critically, the next set of activations also involves signals that go from the feeder back to the chick (following directed links going back from the feeder to the chick through C-B-A and E-D-A). Because the "feeder" is likely to be represented by a node of a high weight (associated with a reinforcement center giving it high value), a strong signal will be sent in the reverse direction on these paths. Additionally, because the link from the feeder to C is relatively weak, node D will receive a stronger signal from the feeder than node B, and consequently the activation level of D will increase more than that of B. Repeated signals of this kind, going from the chick to the feeder and back along the two paths, would result in the activation level of D being ultimately higher than that of B, so that if an execution of the first step of the chick is delayed to this stage, the path A-D-E will be chosen instead of A-B-C. Thus, a simple delay of the execution of the first step caused by setting a higher activation threshold results in a simulation of the two paths (all the way to the food source and back), and thus in a better decision. This is, roughly speaking, how planning works in our framework.

Several points are worth noting from this example: First, we have here another example of the repeated reactivation of nodes and edges, so as to increase their weight and level of activation. We believe that this process is used in many settings. Second, note that we can think of the strength of the signal from B to A as a proxy for the "goodness" of the path through B to the feeder, since it depends on the strength of the signal coming from feeder (which represents the "goodness" or "utility" of the feeder), and the strength of the links from the feeder to C, from C to B, and from B to A, since these depend on the weights of these edges, which is a good proxy for its probability. Similarly, the strength of the signal from D to A is a proxy for the goodness of the path through D to the feeder.

Third, this example makes clear that the threshold for executing actions is an important parameter in the decision-making process of the network. A low threshold results in "impulsive" short-term responses, while a high threshold results in the repeated activation of many potential paths, which can be viewed as "planning via simulation". This threshold is determined by the program of the nodes that execute actions. If the programs of all the nodes in the network are identical, this threshold would be determined by the weight and activation level of that node. But there is no need for all nodes to use the same parameters. The specialization of (the programs of) different nodes by modifying the parameters of their program is possible through evolution, at least for innate nodes, such as those composing the innate stimulus-response (i.e., sensory-effector) components of the network. It is quite likely that the threshold of effector nodes would evolve to be higher precisely because a higher threshold facilitates long-term, less impulsive, planning.

We also expect the extent of planning versus impulsivity to be affected by the environmental setting. When animals are at risk, it makes sense to favor fast responses over long-term planning. The implementation of such a flexible response to risk or urgency may not necessarily require lowering the threshold. It could result from the fact that the level of activation generated by signals such as fear, hunger, and sex drives is high. Strong signals would lead to impulsivity simply because they result in reaching the threshold before a planning simulation can be completed.

Finally, the decision-making process might be improved by making the decision threshold sensitive not to the absolute level of activation, but rather to the relative differences in activation levels between alternative competing paths of actions. The reason is simple: if several paths of actions cross the activation threshold simultaneously or are very similar in their level of activation, the decision may be random or strongly affected by random errors and stochastic fluctuations in signal strength (which is to be expected in a probabilistic system). Here is where further specialization that facilitates competition among signals may be favored. As suggested in Section 2, the program of some nodes can evolve to facilitate the inhibition of one signal by another. When this process is done repeatedly (as suggested above for the simulation process), the path with the consistently higher activation level eventually will be chosen (i.e., the relative difference between its activation level and all others will be above threshold), although this process may take some time.

## 5.4 Relationship to other work on planning and decision making

The approach described so far is quite consistent with current views (and supporting evidence), according to which planning and decision-making are based on memory representation and retrieval (reviewed, for example, by (Weber and Johnson, 2006; Shadlen and Shohamy, 2016; Bellmund *et al.*, 2018)). For example, Weber and Johnson (2006) suggested that the process of constructing predictive utility for making decisions is shaped by three aspects: memory interrogation (the queries posed to memory), accessibility of information, and the structure of memory, which reflects the structure of the world. In our model, the "queries posed to memory" are implemented by the areas of the network that are activated by the sensory nodes upon receiving signals from either the outside world (external stimuli) or from within the body (e.g., hunger or pain signals). The sensory nodes activated, and the strength of their activation level, set off an "activation cascade" in the network, which in our framework is precisely how memory is accessed and retrieved.[8] Thus, in line with Weber and Johnson (2006), slightly different activation profiles of sensory nodes (e.g., due to a stimulus being presented from a slightly different angle or a question being phrased slightly differently) may result in a somewhat different spread of activation in the network, leading to a possibly different decision. Our model also predicts the well-known effects of *priming* (i.e., the increase in the accessibility of memories after previous activation of these memories). These effects, viewed by Weber and Johnson as part of "accessibility of information", emerge naturally in our model. This is because the way that information becomes accessible in our model is through an increase in the activation levels of the nodes representing this information, and because the level of activation is affected by both current and recent activations. As explained earlier, this increase in activation levels is precisely what enables the decision-making process to be based on context-relevant information. Priming in our model is therefore the outcome of a generally adaptive mechanism that raises the activation level of (and thus primes) the parts of the network that are relevant to the current state of the world.

Another phenomenon related to the "accessibility of memory" aspect, also discussed by Weber and Johnson, is *interference* (Bouton, 1993). The activation of some memories reduces the accessibility of other similar or related memories (intuitively, this is the opposite of priming). This competitive effect can emerge in our model as a result of the decision-making process. While, initially, an activation of nodes in our network has the potential to increase rather than decrease the activation level of other nodes, as we pointed out earlier, the decision-making stage can have a competitive nature, and may be affected by the relative level of activation. Thus, when subjects are asked, for example, two related (but not identical) questions, one after the other, the changes in the activation level of nodes and edges caused by the first question will affect the level of activation of some of the nodes and edges used for answering the second question, giving them a relative advantage over other nodes and edges. Dealing with the first question can thus interfere with the decision-making process used to answer the second question. The same will happen with animals that make several decisions sequentially.

---

[8] Note that while the terms "accessibility" and "retrieval" may intuitively imply the existence of a process that accesses and retrieves memories in order to read them or send them elsewhere, we do not have such a separate process in our model. The meaning of "being accessed and retrieved" in our framework is that the nodes and edges that represent memories (that are "accessed") attain a higher level of activation, and that this increase in activation levels affects the activation levels of other nodes and edges in the network.

The third aspect considered by Weber and Johnson in their account of memory-based decision-making is the structure of memory representations. Specifically, they argue that decision-making is affected by how memories are associated with other memories and by the structure of these associations (see also (Kaplan, Schuck, and Doeller, 2017; Kahana 2020)). This aspect clearly plays a central role in our framework. It is captured naturally by the fact that all memories are represented in the network, and that decision making and other behaviors are based on the structure of the network. We hope that this point has become clear to readers by now; the main logic of our model is that the programs run by the nodes and the edges have evolved based on their ability to construct a network that is useful for producing adaptive behavior. This logic is also in line with viewing associative networks as *cognitive maps* in the broad sense, which means that they are used not only for representing the physical space, but also for mapping all statistical relationships in memory representation. Accordingly, planning and thinking can be viewed as navigating a cognitive space (Edelman, 2011; Bellmund *et al.*, 2018).

Two additional aspects of decision-making that our model sheds light on are how value is represented or computed in the brain (Vlaev *et al.*, 2011) and why decision-making takes longer when decisions seem to be more difficult (Shadlen and Shohamy, 2016). Researchers have considered whether decisions are based on absolute values represented in memory or on relative comparisons; both approaches are compatible with our framework. The value of a particular action or item in our network is represented by the weights and the activation levels of all the links and nodes associated with the node representing that particular action or item at a given moment. We might say, for example, that the value of a node representing "choose red feeder" is represented by the weight of the links connecting this choice with "brown grains", of the "brown grains" itself, of the links going from "brown grains" to nodes representing "good tastes", of these specific "good tastes" (not all tastes have the same weight), and of the links associating the "good tastes" nodes with the relevant "reinforcement center". The "red feeder" node may also have a link to a node that represents "empty feeder", which has some negative reinforcement value as a result of having a link of some weight to a node representing "disappointment". While these weights represent the potential value of the red feeder, the decision-making process described above implies that the value is context-specific; it depends on the activation levels of the nodes and edges that determine whether the decision threshold is crossed. It should be clear that the value of the red feeder is very low when a bird is satiated and busy courting a mate or defending a territory; even when it is hungry and looking for food, it will be ranked relative to alternative options. The term "value" is therefore highly state- and process-dependent. All this can be captured by the weights and activation levels of the nodes and links of the network.[9]

Finally, we consider why decision-making takes longer when decisions seem to be more difficult. In a recent review, Shadlen and Shohamy (2016) suggested that "difficult" decisions involves sequential sampling from memory, which takes time because the simultaneous retrieval of memory from many sources is subject to physical bandwidth (or access) constraints (i.e., there is a bottleneck preventing all relevant memories from being retrieved at once). Our model suggests an alternative explanation. As described above, in our model, the decision-making process can be viewed as involving sequential sampling from memory (similar to the process suggested by Shadlen and Shohamy (2016)). Specifically, it involves the repeated activation of areas of the network (by nodes that received internal or external input) until complete paths of actions (i.e., plans) are activated and a threshold for action is crossed. However, in our model there is no bandwidth constraint. A simultaneous activation of all relevant memories is possible simply by using stronger signals, but that would not offer a mechanism for choosing the best action. On the other hand, the process of repeated activation with increasingly stronger signals does provide a mechanism for doing this. When the decision threshold is crossed after only a few activations, we can view the decision as "easy"; the longer it takes to cross the threshold, the more difficult the decision.

# 6. Future challenges: demystifying the hard problems of mind and behavior

To the extent that our theory is "correct" or useful, it should account for (or at least be compatible with) most (if not all) other phenomena and concepts in brain and behavior studies, including the most challenging ones. While showing that this is the case is well beyond the scope of this paper (and, admittedly, beyond our current ability), ignoring these obvious challenges would also not be wise. We therefore end our paper with a few examples of how some (even challenging) concepts may be understood in terms of our network model. We hope that this discussion will encourage future work on these topics.

---

[9] We did not consider here the representation of arithmetic or symbolic values that can also be used for decision making in humans. How such representations are constructed from experience is beyond the scope of this paper

## 6.1 Setting goals and priorities: who is in charge?

If all that we have in the brain is a network, how does the network set goals and priorities for itself? As simple as it might sound, a goal in our framework is any node or a collection of nodes that elicits some actions upon activation. For example, in the very simple case where seeing "brown grains" (and thus activating the "brown grains" node in the network) causes a bird to approach and collect the grains, we can say that "brown grains" were the goal of approaching and collecting. This definition does not require the bird to understand its goal in some conscious way. Observing the activation of nodes and edges by various inputs and the consequences of these activations through planning and execution of plans allows us to determine what are normally viewed as goals.

Recall that most innate responses elicited by stimuli are set by evolution, which means that goals are initially targets of adaptive behavior (e.g., food, a mate, or escape). But once experience affects the structure of the network, any learned representation in the network can potentially become a goal; goals and the actions for achieving them can become quite sophisticated. Still, even an apparently abstract goal such as "writing a novel" has to develop through some association with the biological reward system (e.g., a child may initially learn that speaking, writing, and story-telling are socially and intellectually rewarding, based on innate tendencies to seek social feedback and to find rules and patterns in the environment).

Priorities among goals in our model are determined by the relative level of activation of different goals, just like the decision-making process described earlier. Just as humans and animals need to choose among alternative plans for achieving a goal, they must decide which goal to pursue first. For example, when a male bird decides to forage for food instead of singing, it may reflect the fact that the activation caused by hunger signals was higher than that caused by drivers of mating behavior. This may change quickly if a receptive female is observed nearby, which significantly increases the activation level in the parts of network related to mating behavior.

While this process seems to view humans and animals as simple automata, the process can get much more sophisticated when each potential goal activates a process of long-term planning. We defer more discussion of this topic to future work.

## 6.2 Self-control and free will

We next consider the question of self-control and free will. There is no real "self" within the network – the entire network is the "self". Because the transducers in our framework are probabilistic, even given the same state and input, outcomes and hence network dynamics may develop somewhat differently. Moreover, even in the absence of external stimuli, internal activity may still result in actions. Indeed, it has been suggested that a lack of information from the outside would be programmed to lead to exploratory actions that may generate actions in a semi-random fashion (see (Brembs, 2011)). The logic of this is that in the absence of outside information, a behavior that seeks information is adaptive. Also, in some cases generating a random behavior may be a best response to a predator or a rival (Brembs, 2011).

Following this line, according to our model, the concept of "self-control" is perhaps better viewed as "experience-control" – that is, there is no self that controls the decisions or the actions produced by the network, but long-term planning based on past experiences represented in memory form the "control" mechanism of behavior, preventing wrong or short-cited impulsive decisions (which is what people commonly refer to when discussing self-control).

Free will, like self-control, is merely the "freedom" of the network to use experience in a context dependent way to generate or to inhibit behaviors. The network is "free" to resist outside input or stimuli only to the extent that innate knowledge and past experience represented in the network together with other input, result in such "resistance". Therefore, while in everyday language we might say that resisting temptation is evidence for free will, in terms of our model, it is merely evidence of a more sophisticated process of planning and decision making based on past experience.

Finally, as we have observed repeatedly, the behavior of the network is not deterministic. This adds some freedom to the network, but does not mean that behavior is controlled by a "self". Rather, it means only that the network has the "freedom" (or, perhaps better, the ability) to provide different responses to the same stimuli.

**6.3 Consciousness and self-awareness**

We do not claim to solve here, in a few lines, the complex issue of consciousness (Crick and Koch, 2003; Tononi, 2004; Dehaene, Lau, and Kouider, 2017; Moyal, Fekete, and Edelman, 2020). But we do briefly consider whether there is something in our network model that creates the kind of activities that may be comparable to a "conscious experience". Clearly, at any given moment, there are some portions of the network that are active. The activation levels of nodes and edges reflect what the network has (and has not) "experienced" in the recent past. However, this "experience" of the network would typically not be viewed as "conscious experience". Indeed, what would it even mean for the network to be conscious of its own activity? In other words, what would it take to make the network's activation level and how it changes a conscious experience?

Recall from Section 3.3.2 that the collective activation of nodes and edges at a particular moment forms the working memory that integrates and processes the incoming data. As discussed in Section 3.3.2, some of this collective activity may also be represented in memory as a particular episode in time. Moreover, as described in Section 4.1, the sensory input represented by this activity (of nodes and edges) is matched with pre-existing data in the network, leading to a process that essentially puts the new data in context, generates predictions, influences learning and decision-making, and may also result in the execution of various actions. The network continuously integrates the information it receives, responds to the momentary changes it experienced, and can also represent them as episodes in memory. As a result, new episodes can be compared with previous episodes; each episode activates a range of associations in the network, allowing the activity of the network to effectively position the current episode in context, which is basically how the network "understands" what is going on. We believe that this continuous "understanding" of what is happening "here and now" is not too far from what is normally viewed as consciousness, and is quite compatible with recent computational accounts of this term (Tononi, 2004; Dehaene, Lau, and Kouider, 2017; Moyal, Fekete, and Edelman, 2020).

An additional aspect frequently added to consciousness (but not always – see (Dehaene, Lau, and Kouider, 2017; Moyal, Fekete, and Edelman, 2020)) is that of self-monitoring or self-awareness. In our model, being self-aware requires the network to include some representation of self (i.e., of the agent herself) and to distinguish between data associated with this "self" and data that is associated with others (or with the external world). We have described elsewhere how the construction of the network in our model may support different representations of self and others, and under what conditions the model predicts that this ability would be impaired, as for example in the case of autism (Lotem and Halpern, 2008). We believe that our framework may be particularly helpful in exploring these complexities (see Section 6.5).

## 6.4 Understanding causality

A rat pressing a lever to get a reward and a child discovering that pushing a bottom turns the light on both clearly attain some sense of causality. We can view a directed edge with high weight going from the node representing the action to nodes representing the outcomes as a naïve representation of causality, since (according to the network), the action predicts the outcomes with a high probability. Of course, the edge may just be representing correlation rather than causation. For example, an edge between the light flash due to lightning and the sound of the thunder does not represent a causal relationship. In fact, humans are quite prone to confusing correlation with causality. That said, it is well known that children spend a great deal of effort trying to learn causal relationships (Gopnik *et al.*, 2004). This suggests that we need to be able to represent such relationships in our framework. Indeed, this can be done.

Causal relationships involve *counterfactuals*: Pushing button A causes the door to open because (in part) if were to push a different button A' (counter to fact, since in fact, button A was pushed), the door would not open. Children (and scientists) learn such counterfactuals by experimentation. Such experiments would be represented in our network in a straightforward way. The interventions (e.g., pressing various buttons) would be represented by nodes, as would the outcomes (e.g., the door opening). Since the interventions share a number of features in common, there is likely to be a chunk node, representing a family of related interventions. It then becomes very easy to check whether A causes B: this requires A to (typically) lead to B, and a variant A' of A to be less likely to lead to B. This likely is encoded in the network by the weight of the edges from A to B and from A' to B. (Of course, we are ignoring the many subtleties involves with causality here; see, e.g., (Halpern, 2016) for a discussion. But we believe that this representation should cover the most common cases.)

**6.5 Complex cognitive disorders**

As a high-level theoretical framework, our model cannot address specific neurological impairments. To do that, a more detailed understanding of neuronal mechanisms would be necessary. Yet by focusing on the complex dynamics of the network on which all cognitive abilities are presumably based (i.e., on cognitive development), our model may help to predict what could go wrong. It might also suggest failure modes that could then be compared to real cognitive disorders. For example, our model predicts that the co-evolved coordination between the mechanisms affecting data acquisition and those setting the weight increase and decrease parameters of nodes and edges has critical consequences for the structure of the network (see Section 3.3). Elsewhere, we have suggested that multiple genetic or epigenetic factors that damage the data-acquisition mechanism, especially in the domain of socially mediated data input, can explain many of the cognitive impairments, or cognitive styles, associated with autism (Lotem and Halpern, 2008; Goldstein *et al.*, 2010). Roughly speaking, we suggested that a mismatch between the distribution of data input and the learning parameters (of the type demonstrated in Figure 1B) can result in segmentation problems, leading to atypical network structure with abnormally large chunks. This, in turn, would make generalization and putting incoming data into context more difficult, causing a developmental cascade that we suggested could lead to the symptoms of autism at higher cognitive levels. We believe that our approach might similarly be useful in the analysis of other cognitive disorders that seem to have a high-level complex dynamic, such as schizophrenia or psychosis, as well as in improving our understanding of natural variation in cognitive abilities and intelligence, as well as cognitive biases.

Importantly, the evolutionary approach we took forced us to define the innate components of the network that can be subjected to natural selection.  These are also the components that are likely to be affected by genetic modifications or genetic abnormalities. This implies, for example, that we should be looking, not for genes for "generalization" or "creativity", but rather for the type of genetic modifications that can affect the programs of the nodes and edges that construct the network in a way that can affect the network's ability to generalize or to be creative (Kolodny, Edelman, and Lotem, 2015b).

## Conclusions

We have introduced a new approach for modeling the brain, viewing the brain as a probabilistic transducer that is a network composed of smaller transducers, one associated with each node and edge in the network, all running essentially the same program.  We assume that some initial sections of the network and the parameters of the programs runs by nodes and edges are genetically determined, and could thus be fine-tuned by evolution. The fact that this type of device seems to be able to explain so much of what we expect an autonomous brain to do, using such simple computational devices, suggests that our approach is both computationally and evolutionarily plausible.

What we have provided here is only an outline of a theory that, as we suggested at the beginning of this paper, may help to "decode the magic" of the brain and its evolution, using a set of rules and processes that can in principle do what the brain does. There is clearly much more to be done.  Perhaps most pressing is to test the theory, to see how much it can explain relative to competing theories and the extent to which its predictions are consistent with the evidence. Generating unique predictions of our theory that can be tested experimentally and distinguish it from alternative theories is currently not so easy,  in part because it requires clear alternative theories that are comparable in that they combine mechanistic and evolutionary thinking. Developing such theories and testing predictions that may distinguish between them should certainly be the goal of future work.

Another exciting direction is to combine high-level approaches like ours with some current approaches to related problems.  For example, there has been a great deal of interest recently in the machine-learning community in *one-shot learning*, which, as the name suggests, involves learning (typically, about object categories) from one, or only a few, training examples (see (Wang *et al.*, 2020) for a recent survey). Since in our model we start building our networks based on few examples, trying to combine a one-shot machine-learning approach with our framework may lead both to interesting new insights and to a deeper understanding of both approaches.

In any case, we believe that the time is ripe to develop a computationally and evolutionarily plausible theory of the brain. We hope that our initial steps in this direction provide some impetus to doing that.

# Acknowledgements

We thank Shimon Edelman, Kevin Ellis, David Halpern, Oren Kolodny, Yael Niv, Yosef Prat, Yoav Ram, and Tom Schonberg, for productive as well as critical comments. Of course, none of them is necessarily in full agreement with our views. The first author gratefully acknowledges support from MURI (Multi University Research Initiative) under grant W911NF-19-1-0217, by the ARO under grant W911NF-17-1-0592, by the NSF under grants IIS-1703846 and IIS-1718108, and by a grant from the Open Philanthropy Foundation. The second author gratefully acknowledges support from the Israel Science Foundation (ISF, grants 871/15 and 1126/19).

doi: 10.2139/ssrn.1301075.